\documentclass[sigconf]{acmart}

\usepackage{mathtools}
\usepackage{xcolor}
\usepackage{multirow}
\usepackage{subcaption}
\usepackage{enumitem}
\usepackage{algorithm}
\usepackage{algorithmic}

\theoremstyle{definition}
\newtheorem{definition}{Definition}
\newtheorem{proposition}{Proposition}

\AtBeginDocument{%
  }

\copyrightyear{2023}
\acmYear{2023}
\setcopyright{acmlicensed}\acmConference[WWW '23]{Proceedings of the ACM Web Conference 2023}{May 1--5, 2023}{Austin, TX, USA} \acmBooktitle{Proceedings of the ACM Web Conference 2023 (WWW '23), May 1--5, 2023, Austin, TX, USA}
\acmPrice{15.00}
\acmDOI{10.1145/3543507.3583324}
\acmISBN{978-1-4503-9416-1/23/04}

\begin{document}

\title{Graph Neural Networks with Diverse Spectral Filtering}

\author{Jingwei Guo}
\affiliation{%
    \institution{University of Liverpool}
    \city{Liverpool}
    \country{UK}}
% \affiliation{%
\additionalaffiliation{%
    \institution{Xi’an Jiaotong-Liverpool University}
    \city{Suzhou}
    \country{China}}
\email{Jingwei.Guo@Liverpool.ac.uk}
% \orcid{}

\author{Kaizhu Huang}
\authornote{Corresponding Author}
\affiliation{%
    \institution{Duke Kunshan University}
    \city{Suzhou}
    \country{China}}
\email{Kaizhu.Huang@dukekunshan.edu.cn}

\author{Xinping Yi}
\affiliation{%
    \institution{University of Liverpool}
    \city{Liverpool}
    \country{UK}}
\email{Xinping.Yi@Liverpool.ac.uk}

\author{Rui Zhang}
\affiliation{%
    \institution{Xi’an Jiaotong-Liverpool University}
    \city{Suzhou}
    \country{China}}
\email{Rui.Zhang02@xjtlu.edu.cn}

\renewcommand{\shortauthors}{Jingwei Guo, Kaizhu Huang, Xinping Yi, and Rui Zhang}

\begin{abstract}
Spectral Graph Neural Networks (GNNs) have achieved tremendous success in graph machine learning, with polynomial filters applied for graph convolutions, where all nodes share the \textit{identical} filter weights to mine their local contexts. Despite the success, existing spectral GNNs usually fail to deal with complex networks (e.g., WWW) due to such \textit{homogeneous} spectral filtering setting that ignores the regional \textit{heterogeneity} as typically seen in real-world networks. To tackle this issue, we propose a novel \textit{diverse} spectral filtering~(DSF) framework, which automatically learns node-specific filter weights to exploit the varying local structure properly. Particularly, the diverse filter weights consist of two components --- A global one shared among all nodes, and a local one that varies along network edges to reflect node difference arising from distinct graph parts --- to balance between local and global information. As such, not only can the global graph characteristics be captured, but also the diverse local patterns can be mined with awareness of different node positions. Interestingly, we formulate a novel optimization problem to assist in learning diverse filters, which also enables us to enhance any spectral GNNs with our DSF framework. We showcase the proposed framework on three state-of-the-arts including GPR-GNN, BernNet, and JacobiConv. Extensive experiments over 10 benchmark datasets demonstrate that our framework can consistently boost model performance by up to 4.92\% in node classification tasks, producing diverse filters with enhanced interpretability. 
% added in arXiv version
Code is available at \url{https://github.com/jingweio/DSF}.
\end{abstract}



% \keywords{Graph Neural Networks, Graph Spectral Filtering, Diverse Mixing Patterns, Node Classification}

\maketitle

\section{Introduction}
Recent years have witnessed the explosive growth of learning from graph-structured data. As an emerging technique, Graph Neural Networks (GNNs) have recently attracted significant attention in handling data with complex relationships between entities. Capable of exploiting node features and graph topology simultaneously and adaptively, GNNs achieve state-of-the-art performance in a wide variety of graph analytical tasks, such as social analysis and recommendation system~\cite{Wu2020ACS,chen2020handling}. Spectral GNNs are a class of GNN models that implement convolution operations in the spectral domain. Recent studies show that modern variants mostly function as polynomial spectral filters~\cite{defferrard2016convolutional,kipf2017semi,bianchi2021graph,chien2021adaptive,he2021bernnet,Wang2022HowPA}. Specifically, these filters transform the input source (node features) into a new desired space by selectively attenuating or amplifying its Fourier coefficients induced by the graph Laplacian. Existing efforts either design or learn the polynomial coefficients to simulate different types of filters including  low, band, and/or high-pass. Despite their success, high degree polynomials are necessary as typically required by their expressive power~\cite{chien2021adaptive,he2021bernnet,lingam2021piece} so as to reach high-order neighborhoods. Nevertheless, most spectral GNNs would fail practically due to the overfitting and/or over-squashing problem~\cite{alon2021on,chien2021adaptive,Wang2022HowPA}. They usually end up enforcing \textit{identical} filter weights among nodes, albeit lying in different network areas, to mine their distinct local contexts (see details in Section~\ref{sec:motivation}). Namely, the existing spectral GNN models (including GPR-GNN~\cite{chien2021adaptive}, BernNet~\cite{he2021bernnet}, and JacobiConv~\cite{Wang2022HowPA}) are mostly restricted in the \textit{homogeneous} spectral filtering framework, and focus on the uniform filter weights learning. Such limitation is induced by simplifying diverse regional graph patterns as homogeneous ones at different localities.

% However, real-world networks typically exhibit \textit{heterogeneous} mixing pattern~\cite{balcilar2021breaking},
However, real-world networks typically exhibit \textit{heterogeneous} mixing pattern~\cite{suresh2021breaking},
i.e., different graph parts may possess diverse characteristics (e.g. local assortative level could vary across the graph as illustrated in Figure~\ref{fig:divGraphPattern}). Apparently, GNNs with classic \textit{homogeneous} spectral filtering are inadequate to model the varying regional pattern; this could result in poor interpretability on the micro graph mining which is important in accomplishing node-level tasks. A single shared weight set  tends to pull the model in many opposite directions, which may lead to a biased model that merely captures the most common graph patterns while leaving others not well covered. Ideally, in order to properly mine different local contexts, distinct filter weights might be needed for different nodes. To do so, one may parameterize each node a separate set of trainable filter weights. Unfortunately, this would substantially increase model complexity and cause severe overfitting to local noises, especially in case of large-scale graphs with complex linking patterns (see Section~\ref{sec:ablation}).

\begin{figure}[t]
     \centering
     \begin{subfigure}[b]{0.22\textwidth}
         \centering
         \includegraphics[width=0.9\columnwidth]{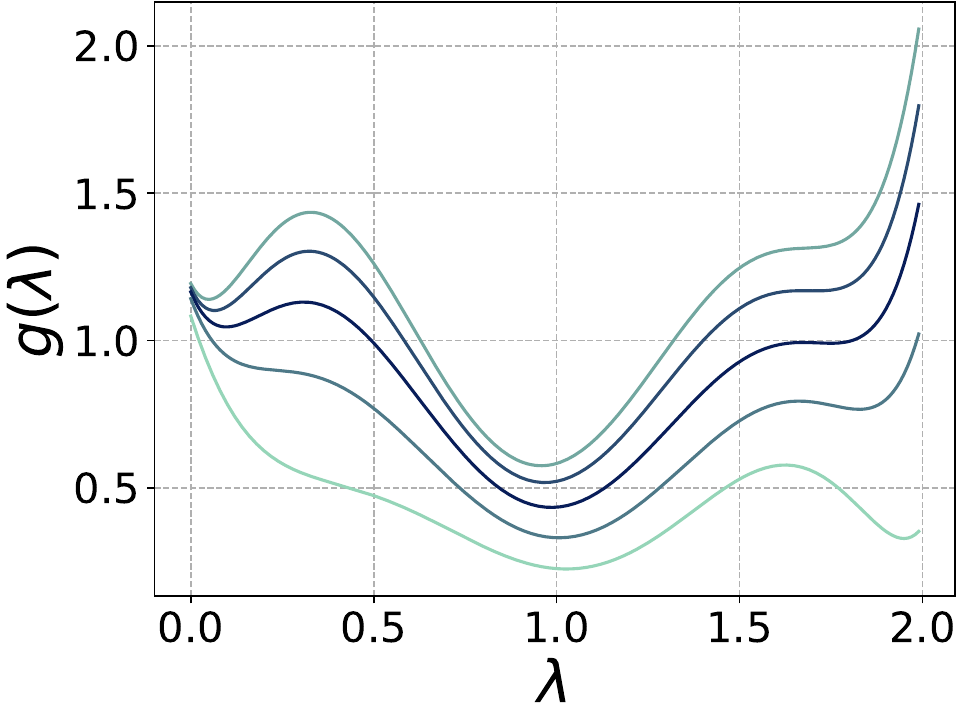}
         \caption{Chameleon}
         \Description{Learned diverse filters on Chameleon dataset.}
         \label{fig:LocFFuns_cham}
    \end{subfigure}
     \begin{subfigure}[b]{0.22\textwidth}
         \centering
         \includegraphics[width=0.9\columnwidth]{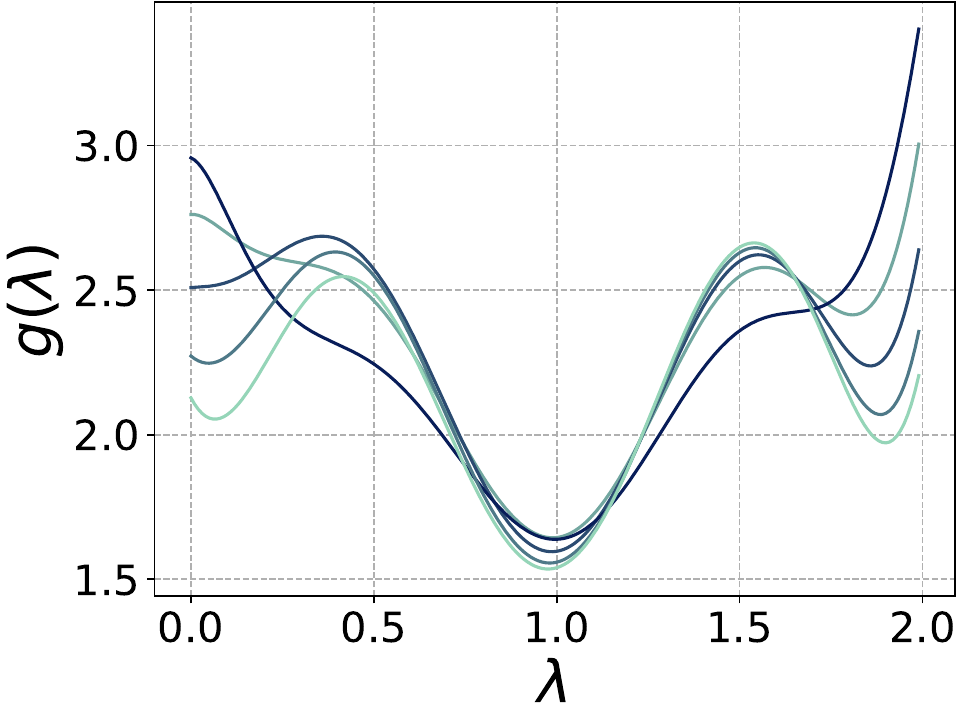}
         \caption{Squirrel}
         \Description{Learned diverse filters on Squirrel dataset.}
         \label{fig:LocFFuns_sq}
     \end{subfigure}
     \begin{subfigure}[b]{0.22\textwidth}
         \centering
         \includegraphics[width=0.9\columnwidth]{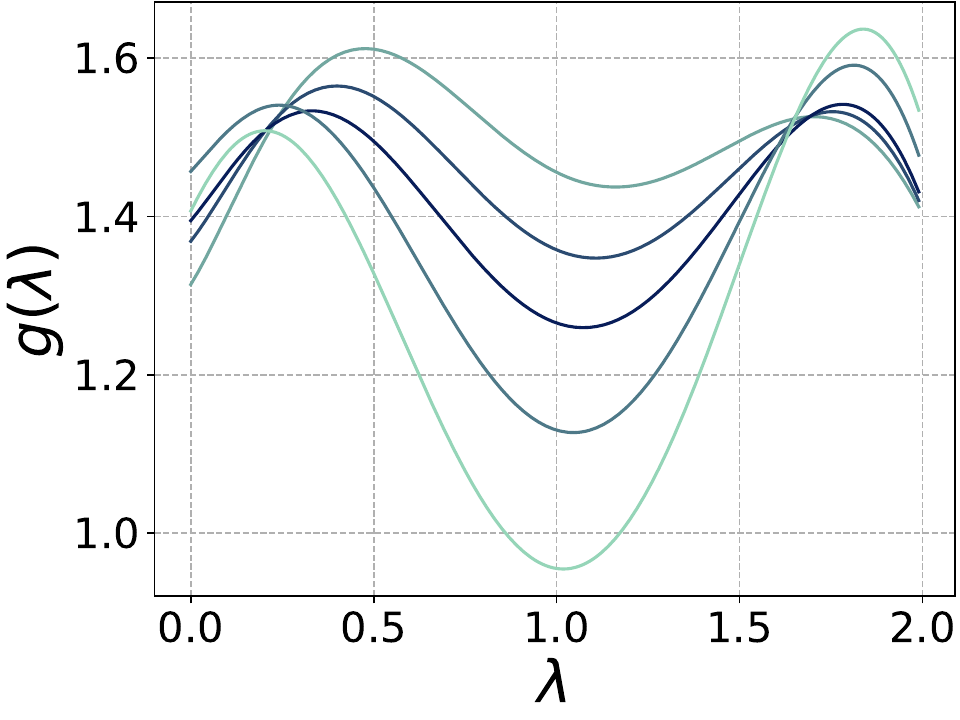}
         \caption{Cornell}
         \Description{Learned diverse filters on Cornell dataset.}
         \label{fig:LocFFuns_cornell}
     \end{subfigure}
    \begin{subfigure}[b]{0.22\textwidth}
        \centering
        \includegraphics[width=1\columnwidth]{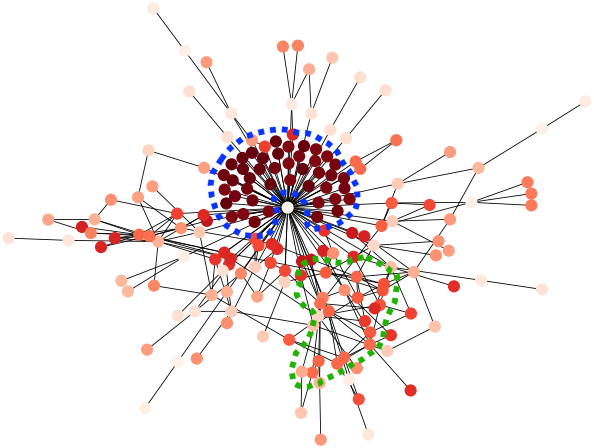}
        \caption{Node-specific Filter Weights}
        \Description{Visualization of node-specific filter weights on Cornell dataset.}
        \label{fig:nxFWs_cornell}    
     \end{subfigure}
\caption{\small{(a)-(c) Diverse filters learned from real-world networks, where five representative curves are plotted for illustration. On each graph, these filters display similar overall shapes but different local details in function curves, showing the capability of our DSF in capturing both the global graph structure and locally varied linking patterns.
(d) Visualization of node-specific filter weights on Cornell dataset, where alike color indicates similar filter weights between nodes. Overall, nodes can be differentiated based on their disjoint underlying regions as circled by the blue and green dashed lines, and far-reaching nodes can still learn similar filter weights due to their akin local structures. E.g., vertices on the graph border are mostly ingrained in a line subgraph such as $\bullet-\bullet-\bullet$, and some unusual cases can be handled (see details in Section~\ref{sec:intep_filter_weight}). These results justify the enhanced model interpretability by learning diverse spectral filters on the micro level.}}
\Description{Figure 1. Fully described in the text.}
\label{fig:intep_visual}
\end{figure}

This work focuses on adapting spectral GNN models to graphs with diverse mixing patterns. We argue that, instead of parameterizing each node arbitrarily different filter weights or a uniform weight, a reasonable design should be \textit{built upon a shared global model whilst locally adapted to each node with awareness of its location in the graph}. Such a proposition is well evidenced by the key observation that nearby nodes tend to display similar local contexts because of their overlapped neighborhoods. For nodes far apart, they are likely to have more possibilities, e.g., even if residing at disjoint graph regions, these nodes may also possess akin local structures due to their similar positions such as graph borders (see Figure~\ref{fig:nxFWs_cornell} and Figure~\ref{fig:nxGraph_cham}). As such, we aim to take advantage of such rationale as a guide to make node-wise adjustments. To this end, we formulate a novel optimization problem to first encode the positional information of nodes. It embeds graph vertices into a low-dimensional coordinate space, based on which we learn node-specific coefficients properly to alter the original filter weights. Accordingly, the global graph filter can be localized on the micro level, allowing individual nodes to adaptively exploit their diverse local contexts. Meanwhile, some beneficial invariant graph properties can still  be preserved by virtue of the mutual filter weights. The proposed framework, named \textit{diverse} spectral filtering~(DSF), flexibly handles the complex graph and makes a proper balance between its conformal and disparate regional patterns with enhanced interpretability, as shown in Figure~\ref{fig:intep_visual}. Besides, our DSF is easy to implement and can be readily plug-and-play in any spectral GNNs with considerable performance gains.

To summarize, the main contributions of this work are three-fold:
\begin{itemize}
    \item We show that many existing spectral GNNs are restricted in the form of \textit{homogeneous} spectral filtering, and identify the need to break this ceiling to deal with complex graphs with regional \textit{heterogeneity}.
    \item We propose a novel \textit{diverse} spectral filtering~(DSF) framework to learn diverse and interpretable spectral filters on the micro level, which consistently leads to performance gains for many spectral GNNs.
    \item We showcase our DSF framework on three state-of-the-arts including GPR-GNN~\cite{chien2021adaptive}, BernNet~\cite{he2021bernnet}, and JacobiConv~\cite{Wang2022HowPA}. Extensive evaluations on 10 real-world datasets demonstrate the superiority of DSF framework in node classification tasks.
\end{itemize}

\section{Related Work}\label{sec:related_work}

\subsection{Spectral Graph Neural Networks}
Recent studies have shown that most spectral GNNs operate as polynomial spectral filters~\cite{chien2021adaptive,he2021bernnet,Wang2022HowPA} with either fixed designs such as GCN~\cite{kipf2017semi}, APPNP~\cite{Klicpera2019PredictTP}, and GNN-LF/HF~\cite{zhu2021interpreting} or learnable forms, e.g., ChebNet~\cite{defferrard2016convolutional}, AdaGNN~\cite{dong2021adagnn}, GPR-GNN~\cite{chien2021adaptive}, ARMA~\cite{bianchi2021graph}, BernNet~\cite{he2021bernnet}, and JacobiConv~\cite{Wang2022HowPA}. 
Further remarks can be found in Appendix~\ref{apdix:related_work_details}.
For both types, it is identified that most spectral GNNs apply a \textit{homogeneous} setting for spectral filtering. These present methods tend to focus on the most frequent graph layouts while under-exploring the rich and diverse local patterns. To alleviate this issue, we introduce a \textit{diverse} spectral filtering (DSF) framework to enhance spectral GNNs with trainable diverse filters. It is worth noting that though both JacobiConv~\cite{Wang2022HowPA} and AdaGNN~\cite{dong2021adagnn} learn multiple filters in a seemingly similar way, they are essentially different from our method in nature. Concretely, their adaptive filters are mainly for studying each feature channel independently, whilist our diverse filters aim at individual context modeling for each node.

\subsection{Graphs with Complex Linking Patterns}
Early wisdom in the community was mainly dedicated to learning from graphs with strong homophiliy~(assortativity) where most connected nodes share similar attributes and same label~\cite{kipf2017semi,velickovic2018graph,Klicpera2019PredictTP}. Until 2020, \citet{pei2020geom} and~\citet{zhu2020beyond} first emphasized the importance of studying GNNs in the heterophily~(disassortativity) setting, thus categorizing real-world networks into homophilic and heterophilic graphs. Recently massive works~\cite{fagcn2021,guo2022gnn,chien2021adaptive,yang2021diverse,yang2022graph} have been done which focus more on complex graph scenarios. For example, FAGCN~\cite{fagcn2021} captured both similarity and dissimilarity between pairwise nodes. As much attention from the community was paid in analyzing graph patterns on the macro/global-level, some researchers~\cite{suresh2021breaking,ma2022meta} start looking into the micro/local graph structures surrounding nodes. In particular, \citet{suresh2021breaking} introduced a node-level assortativity to show heterogeneous mixing patterns inherent in real-world graphs. However, these existing works mostly tackle this regional heterogeneity phenomenon under the intuitive message passing framework~\cite{gilmer2017neural}. Surprisingly, none of them attempts to solve it from the spectral perspective, a theoretically more elegant framework. To this end, we explore in this paper how to learn spectral GNNs on graphs with diverse mixing patterns and propose a novel framework called \textit{diverse} spectral filtering.

It is noted that a recent proposal~\cite{Yang2022PAGNNPG} shares some similarity with our work. This method takes the idea of dynamic neural networks~\cite{han2021dynamic}, and introduce PA-GNN based on GPR-GNN~\cite{chien2021adaptive} by learning node-specific weight offsets. 
Though PA-GNN is also originated in a spectral-based framework, it leverages a simple but less-justified node-specific aggregation scheme and encodes different (but probably inconsistent) sources of information for prediction. On one hand, this would fail to learn interpretable filters; on the other hand, such drawback limits the accuracy gain and may even hurt the performance, which can be later seen in the experiment part.

\subsection{Positional Encoding}
Positional encoding (PE) aims to quantify the global position of, e.g., pixels in images, words in texts, and nodes in graphs. It plays a crucial role in facilitating various neural networks such as CNNs~\cite{islam2020much}, RNNs~\cite{graves2012long}, and Transformer~\cite{dufter2022position}. For GNNs, PE has been widely used to increase their model expression~\cite{bouritsas2022improving,wang2022equivariant,dwivedi2022graph} as bounded by Weisfeiler-Leman (WL) graph isomorphism test~\cite{weisfeiler1968reduction,xu2018how,morris2019weisfeiler}. However, existing focus is mainly put on message-passing GNNs~\cite{gilmer2017neural}, while few research has been conducted on the models defining graph filter in the spectral domain. Recently, \citet{Wang2022HowPA} has established a connection between WL test and the expressive power of spectral GNNs, which motivates us to investigate further how to leverage PE for spectral GNN models. We note that PA-GNN~\cite{Yang2022PAGNNPG} also extract latent positional embeddings from the graph structure, which however cannot be better changed/adjusted to tasks and is mixed with other (even incompatible) attributes. Compared to this, our DSF framework model the positional information of nodes in an independent channel via an iterative updating, thereby producing more expressive and task-beneficial representations. The advantages of this decoupled learning paradigm on node positional and structural features are also verified by other recent works~\cite{dwivedi2022graph,wang2022equivariant}. 

\begin{figure}[t]
     \centering
     \begin{subfigure}[b]{0.4\textwidth}
         \centering
         \includegraphics[width=1\columnwidth]{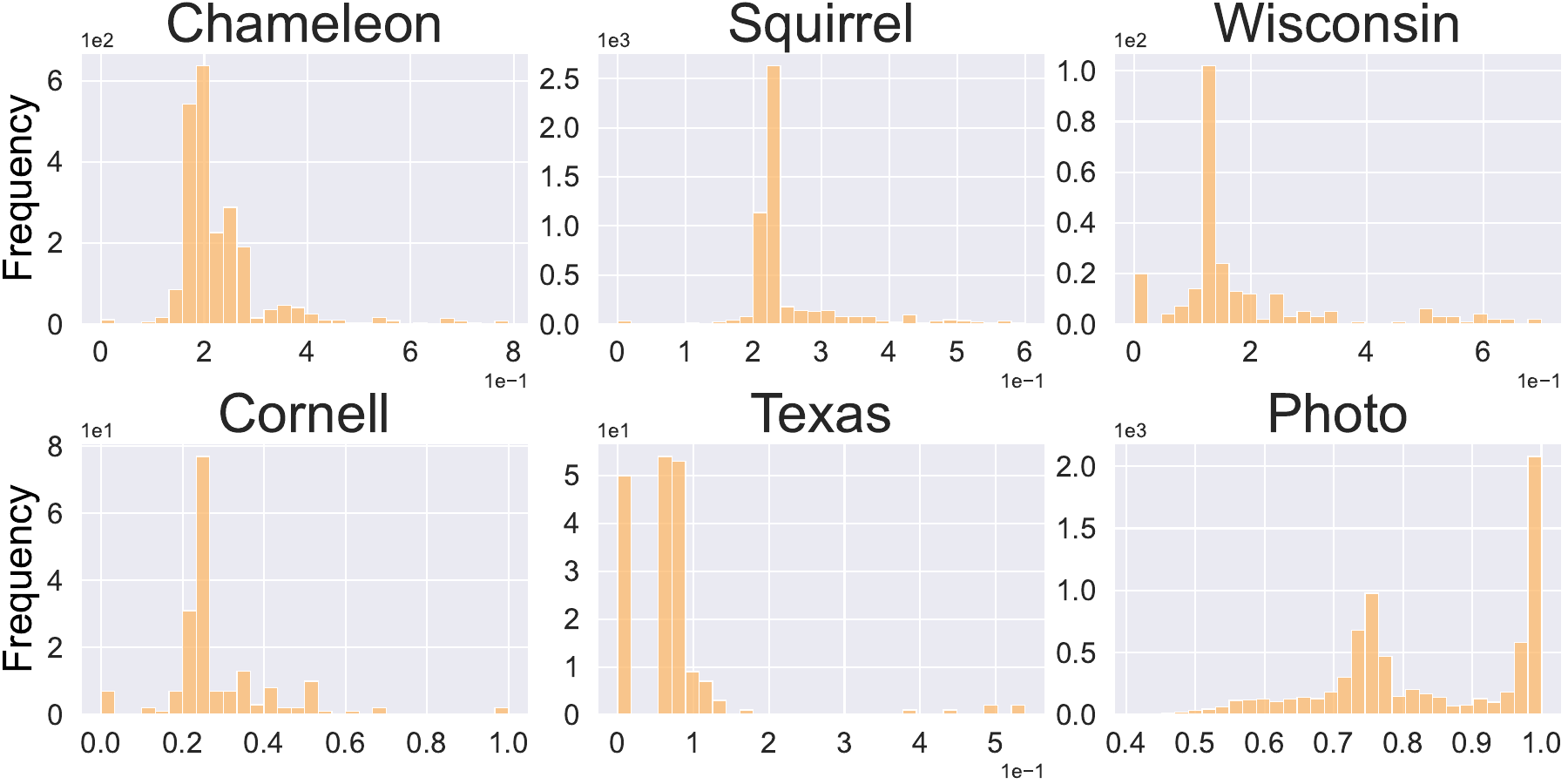}
         \caption{Local Label Homophily}
         \Description{Histograms of Local Label Homophily.}
         \label{fig:Hom_LocDiv}
     \end{subfigure}
     \begin{subfigure}[b]{0.4\textwidth}
         \centering
         \includegraphics[width=1\columnwidth]{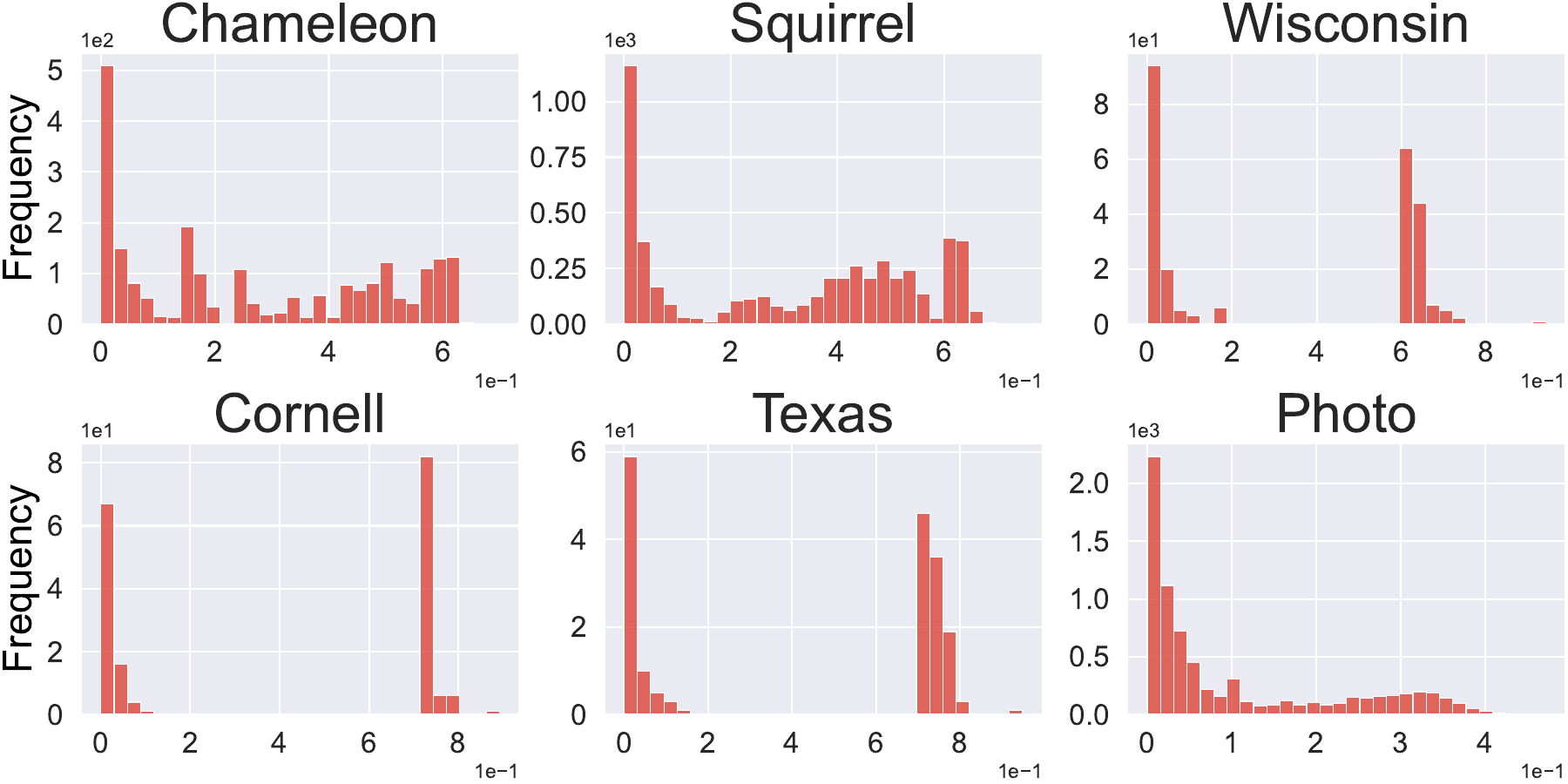}
         \caption{Local Graph Frequency}
         \Description{Histograms of Local Graph Frequency.}
         \label{fig:Ufreq_LocDiv}
     \end{subfigure}
\caption{\small{Distributions of two graph properties on various real graph data (see details in Section~\ref{sec:motivation}).}}
\Description{Distributions of two graph properties on Chameleon, Squirrel, Wisconsin, Cornell, Texas, and Photo datasets.}
\label{fig:divGraphPattern}
\end{figure}

\section{Notations and Preliminaries}\label{sec:prelim}
Let $\mathcal{G} = (\mathcal{V}, \mathcal{E})$ denote an undirected graph with node set $\mathcal{V} = \{v_n\}_{n=1}^N$ with $N=|\mathcal{V}|$ and edge set $\mathcal{E}$ where $(v_i, v_j) \in \mathcal{E}$ if two distinct nodes $v_i$ and $v_j$ are linked by an edge. We use $\mathcal{N}_{i,k} = \{v_j|dis(v_i, v_j) \leq k,\forall v_j \in \mathcal{V}\}$ to represent the $k$-hop neighborhood of node $v_i$. The graph topology is symbolized with an $N \times N$ adjacent matrix $\mathbf{A}$ such that $\mathbf{A}_{i,j}=1$ if $(v_i, v_j) \in \mathcal{E}$ and 0 otherwise. The degree matrix $\mathbf{D}$ is a diagonal matrix with node degrees, e.g., $\text{deg}_i$ w.r.t. node $v_i$, in its diagonal elements. The graph Laplacian matrix is defined as $\mathbf{L} = \mathbf{D} - \mathbf{A}$ with normalized version $\hat{\mathbf{L}}= \mathbf{I} - \hat{\mathbf{A}}$ where $\hat{\mathbf{A}}=\mathbf{D}^{-\frac{1}{2}}\mathbf{A}\mathbf{D}^{-\frac{1}{2}}$ and $\mathbf{I}$ is an $N \times N$ identity matrix. Node features are denoted by $\mathbf{X} \in \mathbb{R}^{N \times f}$ with $f$ being the raw feature dimension. We use $\mathbf{x}_i$ to represent the $i^\text{th}$ row of $\mathbf{X}$ with respect to node $v_i$. In node classification tasks, each node $v_i$ is assigned with a class $c_i$. Thus it has a ground truth one-hot vector $\mathbf{y}_i \in \mathbb{R}^C$, where $C \leq N$ denotes class number. Real-world networks are then divided into homophilic and heterophilic graphs with edge homophily ratio~\cite{zhu2020beyond} on node labels, $\mathcal{H} = \frac{|\{(v_i, v_j) | \mathbf{y}_i = \mathbf{y}_j \wedge (v_i, v_j) \in \mathcal{E} \}|}{|\mathcal{E}|}$ ranging from 0 to 1 with higher values suggesting higher homophily (lower heterophily).

\subsection{Laplacian Decomposition}
Let $\hat{\mathbf{L}} = \mathbf{U} \mathbf{\Lambda} \mathbf{U}^T$ denote the eigen decomposion of $\hat{\mathbf{L}}$, where $\mathbf{U} = [\mathbf{u}_1, \mathbf{u}_2, ..., \mathbf{u}_N] \in \mathbb{R}^{N \times N}$ is a matrix of eigenvectors, and $\mathbf{\Lambda} = \text{diag}(\lambda_1,\lambda_2,...,\lambda_N)$ is the diagonal matrix of eigenvalues. With $\hat{\mathbf{L}}$ being positive semidefinite, we have $\mathbf{U}\mathbf{U}^T=\mathbf{U}^T\mathbf{U}=\mathbf{I}$, of which $\{\mathbf{u}_n\}^N_{n=1}$ is also called the Laplacian eigenbases. We then have
\begin{equation}\label{eq:GloFreq}
\lambda_n = \mathbf{u}_n^T \hat{\mathbf{L}} \mathbf{u}_n = \sum_{(v_p, v_q) \in \mathcal{E}} (\frac{1}{\sqrt{\text{deg}_p}}\mathbf{u}_{n,p} - \frac{1}{\sqrt{\text{deg}_q}} \mathbf{u}_{n,q})^2
\end{equation}
which measures the frequency or smoothness level of each eigenbasis $\mathbf{u}_n$ on the graph. In this paper, we refer to $\lambda_n$ as the global graph frequency. Based on all above, the graph Fourier transform and inverse Fourier transform can be respectively formulated as $\mathbf{S} = \mathscr{F}(\mathbf{X}) = \mathbf{U}^T \mathbf{X}$ and $\mathbf{X} = \mathscr{F}^{-1}(\mathbf{S}) = \mathbf{U} \mathbf{S}$, where $\mathbf{S}$ is often called as the Fourier transformed features or Fourier coefficients of $\mathbf{X}$.

\subsection{Graph Spectral Filtering}
The central idea of spectral GNNs is to transform the graph signal (an instance of node features) in the Fourier space by applying graph spectral filters. They usually take the form as $\mathbf{Z} = g(\hat{\mathbf{L}}) \mathbf{X} = \mathbf{U} g(\mathbf{\Lambda}) \mathbf{U}^T \mathbf{X}$ where $g: [0, 2] \to \mathbb{R}$ is a filter function defined on the spectrum of graph Laplacian. This function creates frequency response to filter different components of $\mathbf{X}$ on Laplacian eigenbases. Take one-channeled $\mathbf{X}$ as an example. As $\mathbf{S} = \mathbf{U}^T \mathbf{X} = [s_1, s_2, ..., s_N]^T$, we have $\mathbf{X}=\sum_{n=1}^N s_n \cdot \mathbf{u}_n$ and $\mathbf{Z} = \sum_{n=1}^N \Bigl( g(\lambda_n) s_n \Bigl) \cdot \mathbf{u}_n$ with scalar $s_n$. It can be seen that node feature matrix $\mathbf{X}$ is mapped into a new $\mathbf{Z}$, by either decreasing or increasing Fourier coefficients $\mathbf{S}$ via $s_n \mapsto g(\lambda_n) s_n$ selectively. Recent studies have  shown that most spectral GNN models implement the filter function $g(\cdot)$ with polynomials~\cite{he2021bernnet,bianchi2021graph,Wang2022HowPA} following $\mathbf{Z} = \sum_{k=0}^{K} \omega_k \hat{\mathbf{L}}^k \mathbf{X} = \sum_{k=0}^{K} \alpha_k P_k(\hat{\mathbf{L}}) \mathbf{X}$, where both $\omega_k$ and $\alpha_k$ denote the polynomial coefficient (also called filter weight), and $P_k: [0,2] \to \mathbb{R}$ is a polynomial basis in the $k^\text{th}$ order. Taking one-channel $\mathbf{X}$ as an example, the existing spectral GNNs can then be unified as
\begin{equation}\label{eq:poly_gsf}
\mathbf{Z} 
= \sum_{k=0}^{K} \alpha_k P_k(\hat{\mathbf{L}}) \mathbf{X}
= \sum_{n=1}^N \hat{s}_n \cdot \mathbf{u}_n
\end{equation}
where $\hat{s}_n=\sum_{k=0}^{K} \alpha_k P_k(\lambda_n) s_n$ for brief symbolization. Present efforts either fix or learn filter weights with different classes of polynomial basis. For instance, APPNP~\cite{Klicpera2019PredictTP} leverages personalized PageRank~\cite{page1999pagerank} to make polynomial basis $P_k(\lambda)=(1 - \lambda)^k$ and set $\alpha_k=\frac{\alpha^k}{1-\alpha}$ with constant hyper-parameter $\alpha$. As an extension of APPNP, GPR-GNN~\cite{chien2021adaptive} directly trains $\alpha_k$ with gradient descent. A comprehensive summarization of various spectral GNNs as polynomial spectral filters can be found in~\cite{Wang2022HowPA}.

\section{diverse spectral filtering}
In this section, the motivation of \textit{diverse} spectral filtering is first provided with both theoretical and empirical analysis. We then present our novel diverse filtering framework.

\subsection{Motivations}\label{sec:motivation}
The unified formula in Eq.~\ref{eq:poly_gsf} can be considered as the \textit{homogeneous} spectral filtering, where all nodes share the identical coefficient $\hat{s}_n$ equally operated on their basis signals, i.e., all elements in $\mathbf{u}_n$, for feature transformation. It seems reasonable as one can learn arbitrary $\hat{s}_n$ with a polynomial graph filter~\cite{shuman2013emerging}, which formally requires high-degree polynomials and reaching high-order node neighborhood~\cite{chien2021adaptive,he2021bernnet,lingam2021piece}. However, aggregating/passing information across a long path via $\hat{\mathbf{L}}^k \mathbf{X}$ with $k \to \infty$ is prone to cause overfitting to noises and/or over-squashing problem~\cite{alon2021on}. \citet{chien2021adaptive} practically show that the polynomial coefficients in Eq.~\ref{eq:poly_gsf} converges to zero as $k$ gets larger. With this empirical finding, \citet{Wang2022HowPA} even propose strategies to optimize $\{\alpha_k\}^K_{k=0}$ with decreasing scale. All the above reveal the local modeling nature of the existing spectral GNNs. In other words, nodes, albeit lying in different graph parts, are enforced to mine their distinct local contexts with the identical filter weights $\{\alpha_k\}^K_{k=0}$. Such filtering scheme implicitly assumes the similar distributions between different graph regions.

This hypothesis however may not be accurate due to the intrinsic complexity in forming real-world networks. To make further investigation, we define two essential graph properties on the local graph level to empirically observe their changing behaviors across the graph. The definitions are given below.

\begin{definition}[Local Label Homophily]\label{def:LocHom}
We define the Local Label Homophily as a measure of the local homophily level surrounding each node $v_i$:
\begin{displaymath}\label{eq:LocHom}
h_i = \frac{|\{(v_p,v_q)|\mathbf{y}_p = \mathbf{y}_q \wedge (v_p,v_q) \in \mathcal{E}_{i,k}\}|}{|\mathcal{E}_{i,k}|}
\end{displaymath}
Here, $h_i$ directly computes the edge homophily ratio~\cite{zhu2020beyond} on the subgraph made up of the k-hop neighbors, and $\mathcal{E}_{i,k} = \{(v_p,v_q) | v_p,v_q \in \mathcal{N}_{i,k} \wedge (v_p,v_q) \in \mathcal{E}\}$ denotes its edge set.
\end{definition}

\begin{definition}[Local Graph Frequency]\label{def:LocUFreq}
The Local Graph Frequency is defined by measuring the local smoothness level of the decomposed Laplacian eigenbases, and for each node $v_i$ we have:
\begin{displaymath}\label{eq:LocUFre}
\lambda_{n,i} = \sum_{(v_p, v_q) \in \mathcal{E}_{i,k}} (\frac{1}{\sqrt{\text{deg}_p}}\mathbf{u}_{n,p} - \frac{1}{\sqrt{\text{deg}_q}} \mathbf{u}_{n,q})^2
\end{displaymath}
where $\lambda_{n,i}$ denotes the frequency or smoothness level of each Laplacian eigenbasis $\mathbf{u}_n$ upon the subgraph induced by the $k$-hop neighbors. Since all summed elements in Eq.~\ref{eq:GloFreq} are positive and $\mathcal{E}_{i,k} \subseteq \mathcal{E}$, we can always have a $\xi_i \in (0, 1)$ such that $\lambda_{n,i} = \xi_i \lambda_n$.
\end{definition}

Figure~\ref{fig:divGraphPattern} shows the distributions of these two graph properties on various real-world data from different domains (see details in Section~\ref{sec:dataset}). We take two-hop neighborhood to illustrate the local graph pattern and remove a few extreme statistical values for better visualization. Moreover, as we usually have a large number of the decomposed eigenbases $\{\mathbf{u}_n\}^N_{n=1}$, it is not quite feasible to visualize them all. Thus, we rank them based on their global graph frequency $\lambda_n$, and calculate the local graph frequencies with the middle one as the most representative demonstration. Similar statistics with low- and high-frequency can be found in the Appendix (see Figure~\ref{fig:supp__divGraphPattern}).

Overall, we observe skewed and even multi-modal distributions. These phenomena imply that the local structural patterns are not uniformly distributed between different graph regions, but exhibiting evident heterogeneity. More importantly, in the spectral domain, the global graph frequency $\lambda_n$ usually fails to capture the diverse local characteristics of $\mathbf{u}_n$ as shown in Figure~\ref{fig:Ufreq_LocDiv}. Thus, weighing $\mathbf{u}_n$ with the only one scalar coefficient, $\hat{s}_n$ computed as a function of $\lambda_n$, may not be appropriate and tends to cause ineffective modeling. In other words, a diverse filtering framework appears necessary so that one could fully exploit the heterogeneous mixing patterns for adaptive micro graph learning.

\subsection{Diverse Filtering Framework}
To implement diverse filtering, we aim to improve the classic \textit{homogeneous} spectral filtering by endowing each node a different set of transforming coefficients on the basis signals. Particularly, the scalar $\hat{s}_n$ is expanded as a vector $\hat{\mathbf{s}}_n=[\hat{\mathbf{s}}_{n,1},\hat{\mathbf{s}}_{n,2}, ..., \hat{\mathbf{s}}_{n,N}]^T$ with the same dimensions as the eigenbasis $\mathbf{u}_n$. Eq.~\ref{eq:poly_gsf} can be thereby enhanced into $\mathbf{Z} = \sum^N_{n=1} \hat{\mathbf{s}}_n \odot \mathbf{u}_n$ where $\odot$ denotes element-wise multiplication, and each element in $\hat{\mathbf{s}}_n$ independently operates on the corresponding signal in $\mathbf{u}_n$. Since $\hat{s}_n$ is originally expressed as a polynomial function of $\lambda_n$, it is reasonable to make 
\begin{equation}\label{eq:locSmth_poly}
\hat{\mathbf{s}}_{n,i} = f(\lambda_{n,i}) =\sum_{k=0}^{K} \alpha_k P_k(\lambda_{n,i}) s_n
\end{equation}
based on our analysis in the previous section, where $\lambda_{n,i}$ denotes the local graph frequency specified in Definition~\ref{def:LocUFreq}. However, it would be computationally expensive to calculate $\lambda_{n,i}$, which requires not only Laplacian decomposition but also subgraph extraction. To mitigate this issue, we turn to exploiting the substitution using $\lambda_{n,i}=\xi_i \lambda_n$ s.t. $0 < \xi_i < 1$ with the following proposition. 

\begin{proposition}\label{prop:poly_subst}
Suppose a K-order polynomial function $f: [0, 2] \to \mathbb{R}$ with polynomial basis $P_k(\cdot)$ and coefficients $\{\alpha_k\}_{k=0}^K$ in real number. For any pair of variables $x,\hat{x} \in [0,2]$ satisfying $x = \xi \hat{x}$ where $\xi$ is a constant real number, we always have a function $g: [0, 2] \to \mathbb{R}$ with the same polynomial basis but a different set of coefficients $\{\beta_k\}_{k=0}^{K}$ such that $f(x) = g(\hat{x})$.
\end{proposition}

Proposition~\ref{prop:poly_subst} suggests that the polynomial $f(\lambda_{n,i})$ computing $\hat{\mathbf{s}}_{n,i}$ in Eq.~\ref{eq:locSmth_poly} can be reformulated into another function of variable $\lambda_n$, using the same basis $P_k(\cdot)$ but a different coefficient set $\{\beta_k\}_{k=0}^{K}$, i.e., $\hat{\mathbf{s}}_{n,i}=f(\lambda_{n,i})=\sum_{k=0}^{K} \alpha_k P_k(\xi_i\lambda_n) s_n=\sum_{k=0}^{K} \beta_{k, i} P_k(\lambda_n) s_n$. Therefore, our \textit{diverse} spectral filtering can be formulated as:
\begin{displaymath}\label{eq:explic_div_filter}
\mathbf{Z} 
= \sum^N_{n=1} \hat{\mathbf{s}}_n \odot \mathbf{u}_n
= \sum_{k=0}^{K}
\text{diag}(\beta_{k, 1}, \beta_{k, 2}, ..., \beta_{k, N})
P_k(\hat{\mathbf{L}}) \mathbf{X}
\end{displaymath}
where each node $v_i$ is parameterized with a different set of filter weights $\{\beta_{k,i}\}_{k=0}^{K}$. The remaining issue is then how to learn these weights. Existing state-of-the-art spectral GNNs~\cite{chien2021adaptive,he2021bernnet,Wang2022HowPA} usually set filter weight as free parameters to be directly trained. However, in our framework, doing this would not only lead to a high computational complexity, but also could cause severe overfitting to local noises. In the following, we introduce two strategies so as to deal with the issue.

\subsubsection{\textbf{Position-aware Filter Weights}}\label{sec:pos_aware_filter_weight}
It has been shown that $\{\beta_{k,i}\}_{k=0}^{K}$ is utilized to mine the local context of each node $v_i$. The differences among these filter weight sets are meant to capture regional heterogeneity on the graph. From another angle, if the filter weights are learned to be similar between nodes, they are more likely to lie in the same region sharing almost identical local structural patterns. 
While, distant node pairs may have more possibilities, e.g., even if residing at disjoint graph regions, these vertices could still possess alike local subgraphs due to their similar positions in the network such as graph borders. This motivates us to make use of node positional information as a guide to learn diverse filter weights.

To do so, the first step is to encode the node positions into a latent space while preserving their graph-based distance. To attain this, inspired by graph signal denoising~\cite{zhu2021interpreting} and Laplacian loss~\cite{belkin2003laplacian,lai2014splitting}, we formulate an novel optimization problem with the objective $\mathcal{L}_p$:
\begin{equation}\label{eq:pos_obj_prob}
\mathop{\arg \min }_{\mathbf{P}}\ \mathcal{L}_p = 
\|\mathbf{X}_p - \mathbf{P}\|^2_F + \kappa_1  tr(\mathbf{P}^T\hat{\mathbf{L}}\mathbf{P}) + 
\kappa_2 \|\mathbf{P}^T \mathbf{P} - \mathbf{I}_d\|^2_F
\end{equation}
where $\mathbf{P} = [\mathbf{P}_1,\mathbf{P}_2,...,\mathbf{P}_N]^T \in \mathbb{R}^{N \times d}$ is a matrix of node positional embeddings, $\mathbf{X}_p$ initializes $\mathbf{P}$ (more information can be seen in Appendix~\ref{apdix:pe_init}), $\mathbf{I}_d$ is an $d \times d$ identity matrix, and both $\kappa_1$ and $\kappa_2$ are non-negative trade-off coefficients. The first term guides $\mathbf{P}$ to be close to $\mathbf{X}_p$, while the middle term enforces  adjacent nodes to stay closer in the positional latent space. A penalty term is lastly appended to ensure orthogonal feature channels for attaining a valid coordinate system. Minimizing $\mathcal{L}_p$ therefore enables a canonical positioning of nodes in the graph. We take an iterative gradient method to solve Eq.~\ref{eq:pos_obj_prob}, and derive the iterative updating rule:
\begin{equation}\label{eq:raw_pe_iter_orth}
\mathbf{P}^{(k+1)} 
= \eta_1 \mathbf{X}_p + (1 - \eta_1) \Bigl(
(1 + \eta_2) \hat{\mathbf{A}} - \eta_2 (\mathbf{P}^{(k)} {\mathbf{P}^{(k)}}^T)
\Bigl) \mathbf{P}^{(k)}
\end{equation}
where $\mathbf{P}^{(0)}=\mathbf{X}_p$, $\eta_1=\frac{1}{1 + \kappa_1 - 2 \kappa_2}$, $\eta_2 = \frac{2 \kappa_2}{\kappa_1 - 2\kappa_2}$, and the stepsize is set as $\frac{\eta_1}{2}$. Both $\eta_1$ and $\eta_2$ are constant hyper-parameters searched from $\{0,0.1,...,1.0\}$ by 0.1, and the case $\eta_1=1.0$ examines the effectiveness of the initial $\mathbf{X}_p$. By iteratively updating $\mathbf{P}^{(k)}$, the objective $\mathcal{L}_P$ can be progressively minimized to solve the optimization problem. In practical training, we need to normalize $\mathbf{P}^{(k)} {\mathbf{P}^{(k)}}^T$ to ensure numerical stability and benefit computational efficiency. Thus, Eq.~\ref{eq:raw_pe_iter_orth} is enhanced into
\begin{equation}\label{eq:pe_iter_orth}
\mathbf{P}^{(k+1)} 
= \eta_1 \mathbf{X}_p + (1 - \eta_1) \Bigl(
(1 + \eta_2) \hat{\mathbf{A}} - \eta_2 \sigma(\mathbf{P}^{(k)} \mathbf{W} {\mathbf{P}^{(k)}}^T)
\Bigl) \mathbf{P}^{(k)}
\end{equation}
where $\sigma$ is a sigmoid function to produce values between 0 to 1, and $\mathbf{W} \in \mathbb{R}^{d \times d}$ is a learnable mapping matrix to improve model capacity. Besides, we also add a tanh activation function between updating steps to make both positive and negative values in the derived coordinate system, i.e., $\mathbf{P}^{(k+1)} \gets \text{Tanh}(\mathbf{P}^{(k+1)})$. We further refer to this process as iterative positional encoding (IPE).

So far, the positional information of nodes can be encoded into a low-dimensional metric space by applying Eq.~\ref{eq:pe_iter_orth} recursively. To learn polynomial filter weights with awareness of node positions, it is empirically found that a simple yet effective non-linear mapping works well:
\begin{equation}\label{eq:pe_to_beta}
\beta_{k,i} = \sigma_p({\mathbf{W}^{(k)}}^T \mathbf{P}_i^{(k)} + \mathbf{b}^{(k)})
\end{equation}
where $\mathbf{W}^{(k)} \in \mathbb{R}^{d}$ and $\mathbf{b}^{(k)} \in \mathbb{R}$ are learnbale parameters for polynomial order $k$, and $\sigma_p$ is an activation function. As such, we manage to train models appropriately with guidance from node positions, while avoiding the possible overfitting induced by parameterizing each node arbitrarily with different filter weights. Moreover, model complexity can also be greatly reduced with the lowered number of trainable parameters from $N \times (K+1)$ to $(d + 1) \times (K+1)$ where $d \ll N$ is feature dimension. Besides, we show our DSF framework, albeit with the simple mapping formula in Eq.~\ref{eq:pe_to_beta}, is able to deal with complex or even unusual graph cases in Section~\ref{sec:intep_filter_weight}.

\subsubsection{\textbf{Local and Global Weight Decomposition}}\label{sec:LGWD}
Though real-world networks exhibit rich and diverse local patterns, the global graph structure still matters, as it encodes some invariant graph properties while simultaneously pruning local noises. Accordingly, we decompose our node-specific filter weights $\beta_{k,i}$ into two independent coefficients $\gamma_i$ and $\theta_{k, i}$ with multiplication, i.e., $\beta_{k,i} = \gamma_i \theta_{k,i}$. We call $\gamma_i \in \mathbb{R}$ the global filter weight responsible for capturing the global graph structure, and name $\theta_{k, i} \in (-1, 1)$ as local filter weight which is learned by the non-linear mapping in Eq.~\ref{eq:pe_to_beta}. As a benefit, the local coefficients can flexibly rescale and/or flip the sign of the global ones to capture node differences, while global connecting patterns can also be mined with diminished noisy information. In this paper, we call this technique Local and Global Weight Decomposition (LGWD). Additionally, we find that JacobiConv~\cite{Wang2022HowPA} also leverages a similar technique called PCD that decomposes the filter weight as $\alpha_k = \pi_k \prod_{s=1}^k \rho_s$ (we replace their symbols to avoid confusions with ours). This design aims to facilitate model training with a decreasing scale on $\{\alpha_k\}_{k=0}^K$ as $k$ grows, and all the nodes still share the same parameter set. In contrast, our method works on individual vertices through disentangling the globally shared and locally varied node coefficients.

\subsection{Overall Algorithm}\label{sec:dsf_implement}
As our DSF framework is independent of any underlying model, it can flexibly improve any spectral GNNs. The overall pipeline of our DSF framework is presented in Appendix~\ref{apdix:model_details}. In practice, we find that the term $\mathbf{P}^{(k)} {\mathbf{P}^{(k)}}^T$ in Eq.~\ref{eq:raw_pe_iter_orth} involves a high computational complexity in $\mathcal{O}(N^2)$, and possibly causes a memory leak while running models on large-scale graphs. To alleviate this, we remove the corresponding term in the objective function, i.e., $\left\|\mathbf{P}^T \mathbf{P} - \mathbf{I}_d\right\|^2_F$ in Eq.~\ref{eq:pos_obj_prob}, and reformulate it into a regularizer:
\begin{equation}\label{eq:orth_reg}
{\mathcal{L}}_{\text{Orth}} = \left\| \hat{\mathbf{P}}^{(K)} \hat{\mathbf{P}}^{(K)} - {\mathbf{I}}_d \right\|^2_F
\end{equation}
where $\hat{\mathbf{P}}^{(K)}$ is normalized from $\mathbf{P}^{(K)}$ such that each column of $\hat{\mathbf{P}}^{(K)}$ has zero mean and one $l_2$ norm. Accordingly, we set $\eta_2=0$ in Eq.~\ref{eq:pe_iter_orth} and have another hyper-parameter $\lambda_\text{Orth}$ called orthogonal regularization parameter. In training, $\mathcal{L}_\text{Orth}$ is penalized with the task loss as $\mathcal{L} = \mathcal{L}_{\text{task}} + \lambda_\text{Orth}\mathcal{L}_{\text{Orth}}$. We further name this variant as DSF-$x$-R where $x$ denotes the backbone GNN, while referring to the original one as DSF-$x$-I. 

\subsubsection{\textbf{Model Analysis}}\label{sec:dsf_analysis}
The proposed DSF framework extends the existing spectral GNNs as
\begin{displaymath}
\sum_{k=0}^{K} \alpha_k P_k(\hat{\mathbf{L}}) \mathbf{X} \to  \sum_{k=0}^{K} \gamma_k 
\text{diag}(\theta_{k, 1}, \theta_{k, 2}, ..., \theta_{k, N})
P_k(\hat{\mathbf{L}}) \mathbf{X}
\end{displaymath}
where a functional space $\{g_i(\cdot) = \sum^{K}_{k=0} \gamma_k \theta_{k,i} P_k(\cdot) | \forall v_i$ $\in \mathcal{V}\}$ made up of diverse filters is derived to enable node-wise learning. Specifically, the underlying graph region of individual node is mined with a different filter function. Existing spectral GNN models mostly learn with one filter function, and thereby can be strengthened with our DSF framework. Besides, in comparison with the advocated interpretability in BernNet~\cite{he2021bernnet}, our DSF framework is able to offer better interpretability  by further differentiating micro graph structures with learned diverse filters (see Section~\ref{sec:intep_filter_weight}).

\subsubsection{\textbf{Time Complexity}}\label{sec:dsf_complexity}
Since our framework requires node-wise computations with positional features, compared to its underlying GNNs, the model complexity is increased by $\mathcal{O}(N (f_p d + 2 K d + 2d + K + 1) + 2 |\mathcal{E}| K d + 2 N^2 K d)$ in our DSF-$x$-I. By the regularization term $\mathcal{L}_\text{Orth}$, we further mange to reduce it by $\mathcal{O}(N^2 K d)$, and introduce our major model named DSF-$x$-R. The average running time is reported on Table~\ref{tab:run_time}. Despite the slightly higher computational overhead, we argue that  DSF framework works still reasonably efficient in practice, especially considering the remarkable performance gains and the enhanced model interpretability (see Section~\ref{sec:node_class_results} and Section~\ref{sec:intep_filter_weight}).

\section{Experiments}\label{sec:experiment}
In this section, we design experiments to answer the following research questions: (\textbf{RQ1}) How effective is our DSF framework in improving state-of-the-art spectral GNNs for node classification? (\textbf{RQ2}) Is there a negative impact on accuracy when each node is parameterized by a separate set of trainable weights? If so, would the proposed strategies, i.e., Position-aware Filter Weights and Local and Global Weight Decomposition (LGWD) take effect in alleviating this? and (\textbf{RQ3}) Could the proposed DSF framework indeed learn diverse and interpretable filters capturing both the common graph structure and regional heterogeneity? 

\begin{table}[t]
\caption{\small{Average running time per epoch (ms)/average total running time (s). Although DSF-GPR-I is less efficient on large networks, DSF-GPR-R, (our major model) can reduce it by more than 75\% on average (though reasonably slower than GPR-GNN).}}
\Description{Average running time per epoch and average total running time are respectively reported on large-scale and small-scale graphs, by comparing models GPR-GNN, DSF-GPR-I, and DSF-GPR-R.}
\label{tab:run_time}
\setlength\tabcolsep{5pt}
\renewcommand{\arraystretch}{0.85}
\resizebox{0.4\textwidth}{!}{
\begin{tabular}{lcc|c}
\toprule[0.3pt]
\textbf{Datasets} 
&\textbf{Small-scale}      &\textbf{Large-scale}           &\textbf{Average}
\\
\midrule[0.3pt]
GPR-GNN
&1.10/2.24    &0.98/5.01   &1.08/2.74\\
DSF-GPR-I
&5.96/12.19     &40.34/131.77    &12.21/33.93\\
DSF-GPR-R
&2.49/6.29 &3.02/14.48      &2.59/7.78\\
\bottomrule[0.3pt]
\end{tabular}
}
\end{table}

\begin{table*}[t]
\caption{\small{Node classification accuracies (\%) $\pm$ 95\% confidence interval over 100 runs. 
The row of PA-GNN~\cite{Yang2022PAGNNPG}$^*$ lists the relative improvements of PA-GNN upon GPR-GNN based on the results obtained from its paper, where -- denotes values not provided. Our Improv. gives the best relative improvements between our DSF variants over their common underlying model.}}
\Description{Table 2. Fully described in the text.}
\label{tab:real_nc}
\setlength\tabcolsep{2.5pt}
\renewcommand{\arraystretch}{0.87}
\resizebox{0.98\textwidth}{!}{
\begin{tabular}{p{1.9cm}cccccc|ccccc}
\toprule[0.3pt]
% \hline
\multirow{2}{*}{\textbf{Datasets}} & \multicolumn{6}{c}{\textbf{Heterophilic Graphs}}                                          & \multicolumn{5}{|c}{\textbf{Homophilic Graphs}}
% \\ \cline{2-11}
\\ \cmidrule[0.3pt]{2-12}
&\textbf{Chameleon}      &\textbf{Squirrel}                  &\textbf{Wisconsin}     &\textbf{Cornell}      &\textbf{Texas}      &\textbf{Twitch-DE}  &\textbf{Cora}      &\textbf{Citeseer}           &\textbf{Pubmed} &\textbf{Computers}      &\textbf{Photo}  \\
\midrule[0.3pt]
GCN~\cite{kipf2017semi}       &67.22\footnotesize{$\pm$0.43}       &54.21\footnotesize{$\pm$0.41}                     &59.45\footnotesize{$\pm$0.72}       &52.76\footnotesize{$\pm$1.17} &61.66\footnotesize{$\pm$0.71}       
&73.94\footnotesize{$\pm$0.15}
&88.13\footnotesize{$\pm$0.25}       &77.00\footnotesize{$\pm$0.27}             
&89.07\footnotesize{$\pm$0.11}
&91.06\footnotesize{$\pm$0.12}       &93.99\footnotesize{$\pm$0.12} 
 \\
GAT~\cite{velickovic2018graph}       &67.72\footnotesize{$\pm$0.41}       &52.26\footnotesize{$\pm$0.58}                     &57.94\footnotesize{$\pm$0.89}       &50.20\footnotesize{$\pm$0.93} &55.37\footnotesize{$\pm$1.10}      
&73.00\footnotesize{$\pm$0.15}
&88.47\footnotesize{$\pm$0.22}       &77.23\footnotesize{$\pm$0.27}              &88.30\footnotesize{$\pm$0.11}
&91.69\footnotesize{$\pm$0.11}       &94.55\footnotesize{$\pm$0.11} 
\\
ChebNet~\cite{defferrard2016convolutional}       &64.85\footnotesize{$\pm$0.44}       &48.14\footnotesize{$\pm$0.33}                    &80.93\footnotesize{$\pm$0.72}       &77.98\footnotesize{$\pm$1.00}       &75.83\footnotesize{$\pm$1.20}  
&73.73\footnotesize{$\pm$0.14}
&87.64\footnotesize{$\pm$0.21}       &76.93\footnotesize{$\pm$0.24}                 &89.91\footnotesize{$\pm$0.11}
&91.65\footnotesize{$\pm$0.12}       &95.27\footnotesize{$\pm$0.07} 
\\
APPNP~\cite{Klicpera2019PredictTP}       &53.66\footnotesize{$\pm$0.33}       &36.08\footnotesize{$\pm$0.36}                     &81.23\footnotesize{$\pm$0.64}       &81.29\footnotesize{$\pm$0.78} &79.42\footnotesize{$\pm$1.05}    
&72.65\footnotesize{$\pm$0.11}
&88.70\footnotesize{$\pm$0.21}       &77.77\footnotesize{$\pm$0.24}              
&89.93\footnotesize{$\pm$0.09}
&91.62\footnotesize{$\pm$0.10}       &94.92\footnotesize{$\pm$0.09} 
\\
GNN-LF~\cite{zhu2021interpreting}       &54.29\footnotesize{$\pm$0.36}       &36.87\footnotesize{$\pm$0.33}       &59.85\footnotesize{$\pm$0.60}       &62.90\footnotesize{$\pm$0.98}       &61.88\footnotesize{$\pm$0.95}       &73.03\footnotesize{$\pm$0.13}       &88.90\footnotesize{$\pm$0.25}       &77.35\footnotesize{$\pm$0.29}       
&88.89\footnotesize{$\pm$0.10}
&91.12\footnotesize{$\pm$0.11}       &95.13\footnotesize{$\pm$0.08} 
\\
GNN-HF~\cite{zhu2021interpreting}       &55.22\footnotesize{$\pm$0.42}       &35.45\footnotesize{$\pm$0.30}       &68.17\footnotesize{$\pm$0.72}       &72.98\footnotesize{$\pm$1.02}       &66.66\footnotesize{$\pm$1.34}       &71.92\footnotesize{$\pm$0.13}       &89.01\footnotesize{$\pm$0.19}       &77.74\footnotesize{$\pm$0.23}       
&89.53\footnotesize{$\pm$0.10}
&90.73\footnotesize{$\pm$0.10}       &95.26\footnotesize{$\pm$0.09} \\
FAGCN~\cite{fagcn2021}       &68.38\footnotesize{$\pm$0.51}       &50.08\footnotesize{$\pm$0.60}                     &82.11\footnotesize{$\pm$0.85}       &79.00\footnotesize{$\pm$0.93}         &81.00\footnotesize{$\pm$0.95}   
&74.15\footnotesize{$\pm$0.13}
&88.82\footnotesize{$\pm$0.20}       &77.65\footnotesize{$\pm$0.29}              
&90.13\footnotesize{$\pm$0.11}
&91.90\footnotesize{$\pm$0.11}       &95.25\footnotesize{$\pm$0.10} 
\\
% \hline
\midrule[0.3pt]
GPR-GNN~\cite{chien2021adaptive}       &69.01\footnotesize{$\pm$0.50}       &55.39\footnotesize{$\pm$0.33}                     &82.72\footnotesize{$\pm$0.85}       &80.81\footnotesize{$\pm$0.78}         &81.66\footnotesize{$\pm$1.02} 
&74.07\footnotesize{$\pm$0.18}
&89.03\footnotesize{$\pm$0.20}       &77.63\footnotesize{$\pm$0.28}              
&90.10\footnotesize{$\pm$0.44}
&92.34\footnotesize{$\pm$0.13}       &95.34\footnotesize{$\pm$0.09} 
\\
DSF-GPR-I   &71.18\footnotesize{$\pm$0.52}       &57.08\footnotesize{$\pm$0.29}       &\textbf{87.64}\footnotesize{$\pm$0.79}       &84.76\footnotesize{$\pm$0.90}       &85.44\footnotesize{$\pm$1.05}       &74.58\footnotesize{$\pm$0.16}       &\textbf{89.64}\footnotesize{$\pm$0.20}       &78.03\footnotesize{$\pm$0.26}       &90.26\footnotesize{$\pm$0.08}
&92.49\footnotesize{$\pm$0.12}       &95.64\footnotesize{$\pm$0.07}
\\
DSF-GPR-R       &\textbf{71.64}\footnotesize{$\pm$0.55}       &\textbf{58.44}\footnotesize{$\pm$0.30}                     &87.43\footnotesize{$\pm$0.74}       &\textbf{84.93}\footnotesize{$\pm$0.90}       &\textbf{85.56}\footnotesize{$\pm$0.93}
&\textbf{74.81}\footnotesize{$\pm$0.14}
&89.63\footnotesize{$\pm$0.17}       &\textbf{78.22}\footnotesize{$\pm$0.29}       
&\textbf{90.51}\footnotesize{$\pm$0.07}
&\textbf{92.80}\footnotesize{$\pm$0.12}       &\textbf{95.73}\footnotesize{$\pm$0.08} 
\\
Our Improv.         &2.63\%         &3.05\%         &4.92\%         &4.12\%         &3.9\%         &0.74\%         &0.61\%         &0.59\%         &0.41\%          &0.46\%         &0.39\%       \\
PA-GNN~\cite{Yang2022PAGNNPG}$^*$         &0.66\%         &1.28\%         &--         &--         &--         &--         &-0.09\%         &-0.74\%         &-0.03\%        &1.03\%         &0.02\%   \\
% \hline
\midrule[0.3pt]
BernNet~\cite{he2021bernnet}       &70.59\footnotesize{$\pm$0.42}       &56.63\footnotesize{$\pm$0.32}                     &85.00\footnotesize{$\pm$0.94}       &82.10\footnotesize{$\pm$0.95}         &82.20\footnotesize{$\pm$0.98}  
&74.45\footnotesize{$\pm$0.15}
&88.72\footnotesize{$\pm$0.23}       &77.52\footnotesize{$\pm$0.29}              
&90.21\footnotesize{$\pm$0.46}
&92.57\footnotesize{$\pm$0.10}       &95.42\footnotesize{$\pm$0.08} \\
DSF-Bern-I   &72.95\footnotesize{$\pm$0.53}       &59.45\footnotesize{$\pm$0.32}       &\textbf{88.23}\footnotesize{$\pm$0.81}       &\textbf{85.07}\footnotesize{$\pm$0.93}       &\textbf{84.59}\footnotesize{$\pm$1.07}       &74.96\footnotesize{$\pm$0.15}       &89.05\footnotesize{$\pm$0.22}       &\textbf{78.32}\footnotesize{$\pm$0.27}       
&90.40\footnotesize{$\pm$0.10}
&92.76\footnotesize{$\pm$0.10}       &95.73\footnotesize{$\pm$0.07}
\\
DSF-Bern-R       &\textbf{73.60}\footnotesize{$\pm$0.53}       &\textbf{59.99}\footnotesize{$\pm$0.30}                     &88.02\footnotesize{$\pm$0.91}       &84.29\footnotesize{$\pm$0.93}      &84.42\footnotesize{$\pm$1.00}      
&\textbf{75.00}\footnotesize{$\pm$0.15}
&\textbf{89.10}\footnotesize{$\pm$0.22}       &78.27\footnotesize{$\pm$0.26}       
&\textbf{90.52}\footnotesize{$\pm$0.10}
&\textbf{92.84}\footnotesize{$\pm$0.10}       &\textbf{95.79}\footnotesize{$\pm$0.06}
\\
Our Improv.         &3.01\%         &3.36\%         &3.23\%         &2.97\%         &2.39\%         &0.55\%         &0.38\%         &0.80\%         &0.31\%        &0.27\%         &0.37\%      \\
% \hline
\midrule[0.3pt]
JacobiConv~\cite{Wang2022HowPA}       &73.71\footnotesize{$\pm$0.42}       &57.22\footnotesize{$\pm$0.24}                     &83.21\footnotesize{$\pm$0.68}       &82.34\footnotesize{$\pm$0.88}         &82.42\footnotesize{$\pm$0.90}
&74.34\footnotesize{$\pm$0.12}
&89.24\footnotesize{$\pm$0.19}       &77.81\footnotesize{$\pm$0.29}       
&89.50\footnotesize{$\pm$0.47}
&92.26\footnotesize{$\pm$0.10}       &95.62\footnotesize{$\pm$0.06} \\
DSF-Jacobi-I   &74.88\footnotesize{$\pm$0.39}       &58.26\footnotesize{$\pm$0.26}       &85.34\footnotesize{$\pm$0.74}       &\textbf{84.54}\footnotesize{$\pm$0.81}       &83.68\footnotesize{$\pm$1.12}       &74.65\footnotesize{$\pm$0.13}       &89.54\footnotesize{$\pm$0.19}       &78.18\footnotesize{$\pm$0.26}       
&89.78\footnotesize{$\pm$0.09}
&92.38\footnotesize{$\pm$0.11}       &\textbf{95.76}\footnotesize{$\pm$0.07}
\\
DSF-Jacobi-R       &\textbf{75.00}\footnotesize{$\pm$0.38}       &\textbf{59.23}\footnotesize{$\pm$0.27}                     &\textbf{86.13}\footnotesize{$\pm$0.70}       &84.39\footnotesize{$\pm$0.88}        &\textbf{84.46}\footnotesize{$\pm$0.81}
&\textbf{74.75}\footnotesize{$\pm$0.15}
&\textbf{89.66}\footnotesize{$\pm$0.19}       &\textbf{78.23}\footnotesize{$\pm$0.25}       
&\textbf{90.07}\footnotesize{$\pm$0.10}
&\textbf{92.44}\footnotesize{$\pm$0.11}        &95.75\footnotesize{$\pm$0.08}\\
Our Improv.         &1.29\%         &2.01\%         &2.92\%         &2.20\%         &2.04\%         &0.41\%         &0.42\%         &0.42\%     &0.41\%          &0.18\%         &0.14\%    \\
\bottomrule[0.3pt]
% \hline
\end{tabular}}
\end{table*}

\subsection{Datasets and Experimental Setup}\label{sec:data_and_expsetup}

\subsubsection{\textbf{Datasets}}\label{sec:dataset}
We examine models over 11 real-world datasets from various domains including 6 heterophilic graphs as Chameleon, Squirrel~\cite{rozemberczki2021multi}, Wisconsin, Cornell, Texas~\cite{pei2020geom} (webpage networks), and Twitch-DE~\cite{rozemberczki2021multi,lim2021new} (social network), as well as 5 homophilic graphs, i.e., Cora, Citeseer, Pubmed~\cite{sen2008collective} (citation networks), Computers, and Photo~\cite{mcauley2015image,shchur2018pitfalls} (the Amazon co-purchase graphs). Detailed statistics are provided in the Appendix (see Table~\ref{tab:data_sta}). We divide each dataset into 60\%/20\%/20\% for training/validation/testing by following~\cite{chien2021adaptive,he2021bernnet,Wang2022HowPA}, and create 10 random splits for evaluation.

\subsubsection{\textbf{Baselines}}
To verify the effectiveness of the proposed DSF framework, we implement it upon three state-of-the-art spectral GNNs with trainable polynomial filters, i.e., GPR-GNN~\cite{chien2021adaptive}, BernNet~\cite{he2021bernnet}, and JacobiConv~\cite{Wang2022HowPA}. Implementation details are provided in Appendix~\ref{apdix:model_details}. Therefore, we have six variants with names formatted as DSF-$x$-$\phi$ for $x \in \{\text{GPR}, \text{Bern}, \text{Jacobi}\}$ and $\phi \in \{\text{I}, \text{R}\}$. For a more comprehensive comparison, we also consider another 8 baseline GNNs including GCN~\cite{kipf2017semi}, GAT~\cite{velickovic2018graph}, ChebNet~\cite{defferrard2016convolutional}, APPNP~\cite{Klicpera2019PredictTP}, GNN-LF/HF~\cite{zhu2021interpreting}, FAGCN~\cite{fagcn2021}, and PA-GNN~\cite{Yang2022PAGNNPG}.

\subsubsection{\textbf{Setup}}
We fix the number of hidden features $d=64$ for all models, and set the polynomial order $K=10$ to follow~\cite{chien2021adaptive,he2021bernnet,Wang2022HowPA}. For each dataset, we tune the hyper-parameters of all models, including baselines with their specified parameter ranges, on the validation split using Optuna~\cite{akiba2019optuna} for 200 trails. With the best hyper-parameters, we train models in 1,000 epochs using early-stopping strategy and a patience of 100 epochs. The average performance over 100 runs (10 runs $\times$ 10 splits) are reported. For reproducibity, our implementation and the searching space of hyper-parameters are available at \url{https://github.com/jingweio/DSF}.

\subsection{RQ1. Overall Evaluation}\label{sec:node_class_results}
To answer RQ1, we report the average node classification accuracies with a 95\% confidence interval. From Table~\ref{tab:real_nc}, we have the following observations: 
\textbf{1)} Spectral GNNs with trainable filters generally yield better classification results than other baseline models. This is because conventional GNNs typically fail to deal with complex linking patterns, e.g., in heterophilic graphs, using their fixed frequency response filters. Contrastively, GPR-GNN, BernNet, and JacobiConv can simulate different types of filters to learn from both assortative and disassortative label patterns. FAGCN is able to capture both low- and high-frequency information but is limited as they can only model pairwise node relationship.
\textbf{2)} The proposed DSF framework consistently produces performance boost over its underlying models, especially on heterophilic graphs with the maximal improvement up to 4.92\%. This can be mainly explained by the diverse characteristics inherent in their local graph patterns, as shown in Figure~\ref{fig:divGraphPattern}. By comparison, the existing spectral GNN models assume the \textit{homogeneous} spectral filtering, and neglect regional \textit{heterogeneity} at different graph localities.
\textbf{3)} For Twitch-DE dataset, albeit also being heterophilic graphs, we only observe an marginal improvement made by our framework. This is due to the fact that its local structural patterns are naturally similar and distributed uniformly across the graph, as indicated by the concentrated histogram in the Appendix (see Figure~\ref{fig:supp__divGraphPattern}). Similar results can be spotted on homophilic graphs with assortative linking patterns, which can be easily modeled with classic \textit{homogeneous} spectral filtering. The refined classification accuracies made by DSF is mainly contributed by dealing with some sudden changes on the graph boundaries, e.g., between different community regions or social circles. 
\textbf{4)} We conduct comparisons with PA-GNN~\cite{Yang2022PAGNNPG} which also tries to learn node-specific parameter offsets. 
As no codes are publicly available  for PA-GNN, we simply compute the relative improvements upon its base model GPR-GNN by copying the values  from their paper (listed  in the row of PA-GNN~\cite{Yang2022PAGNNPG}$^*$). In general, PA-GNN shows marginal or even negative performance gains. We conjecture that PA-GNN might encode different sources of information  for predicting the offsets with small value constraint, thus limiting their performance.
\textbf{5)} Interestingly, it is noted our variant DSF-$x$-R not only decreases the model complexity of DSF-$x$-I but also achieves higher performance gains on average. This is partially because DSF-$x$-I minimizes the orthogonal penalty, i.e., the last term in Eq.~\ref{eq:pos_obj_prob}, mainly by means of the iterative aggregation on a non-sparse graph computed by $\mathbf{P}^{(k)} \mathbf{W} {\mathbf{P}^{(k)}}^T$. Despite the theoretical convergence, aggregating features on  dense graph is prone to mistakenly preserve noises and cause model overfitting. On the other hand, DSF-$x$-R with the regularization loss $\mathcal{L}_\text{Orth}$ offers a more flexible and accurate control, provides extra supervisory signals directly operated on model parameters, and could also benefit from the advanced optimization technique such as Adam~\cite{kingma2014adam} algorithm.

\begin{table}[t]
\caption{\small{Reduced classification accuracies (\%) of our DSF framework compared to base models while learning without IPE.}}
\Description{Ablation study of the proposed IPE on Chameleon, Squirrel, Wisconsin, Cornell, Texas, and Photo datasets.}
\label{tab:abla_pe}
\setlength\tabcolsep{3pt}
\renewcommand{\arraystretch}{0.85}
\resizebox{0.46\textwidth}{!}{
\begin{tabular}{lcccccc}
\toprule[0.3pt]
\textbf{Datasets} & \textbf{Chameleon} & \textbf{Squirrel} & \textbf{Wisconsin} & \textbf{Cornell} & \textbf{Texas} & \textbf{Photo} \\
\midrule[0.3pt]
DSF-GPR w/o IPE   
&22.62   &22.55   &5.81   &11.20   &11.10   &1.51   \\
DSF-Bern w/o IPE   
&17.47   &18.65   &4.72   &6.25   &6.54   &3.59   \\
DSF-Jacobi w/o IPE  
&24.64   &26.93   &1.10   &3.10   &5.88   &1.53   \\
\midrule[0.3pt]
Average Reduction
&\textbf{21.58}   &\textbf{22.71}   &\textbf{3.88}   &\textbf{6.85}   &\textbf{7.84}   &\textbf{2.21}   \\
\bottomrule[0.3pt]
\end{tabular}}
\end{table}

\subsection{RQ2. Ablation Study}\label{sec:ablation}
\begin{figure}[t]
\centering
\includegraphics[width=0.4\textwidth]{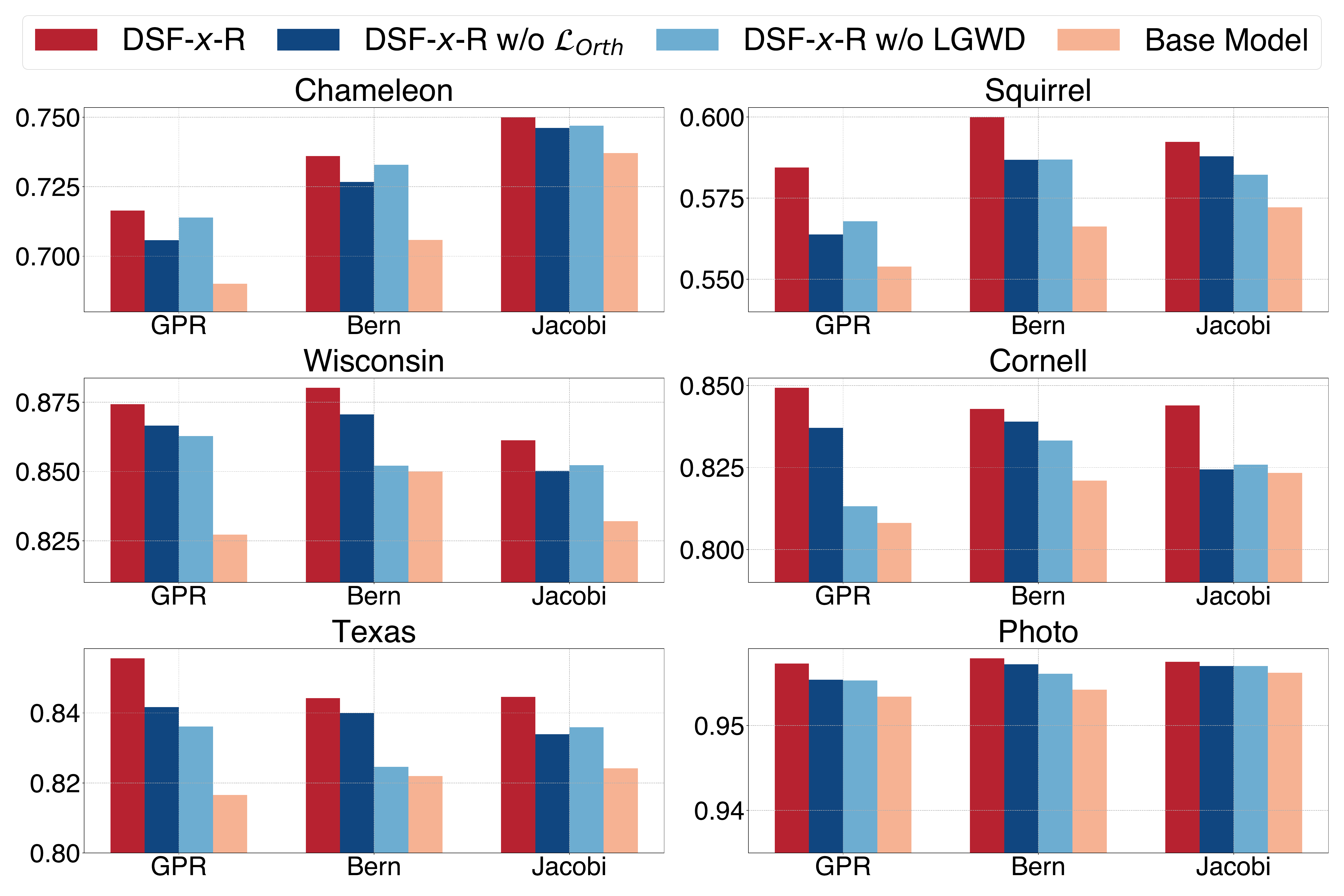}
\caption{\small{Ablation study of DSF framework on six datasets with our variants DSF-$x$-R for all $x \in \{\text{GPR}, \text{Bern}, \text{Jacobi}\}$ as an example.}}
\Description{Performance comparison among our model, its variants while ablating the regularization term and the Local and Global Weight Decomposition technique, as well as the base model.}
\label{fig:abla_study}
\end{figure}

This subsection aims to validate our designs through ablation study. We earlier argue that it is inappropriate to directly parameterize each node a separate set of trainable filter weights. To provide empirical evidences, we first experiment with our DSF framework in node classification tasks while ablating the module of iterative positional encoding (IPE). That is to directly make $N \times (K+1)$ filter weights w.r.t. nodes to be trained as model parameters. We then report the downgraded model performance compared to the underlying models in Table~\ref{tab:abla_pe}, where six datasets are experimented for illustration. As observed, learning without IPE leads to a clear accuracy drop, notably on networks Chameleon and Squirrel with complex connecting patterns and relative a large number of nodes. This confirms our early conjecture as well as the importance of the proposed IPE strategy. In this work, we also constrain the channel orthogonality while encoding positional features, and introduce a technique called Local and Global Weight Decomposition (LGWD). To examine their effectiveness, we conduct comprehensive ablation study over six datasets in node classification. For simplicity, we take DSF-$x$-R, one variant of our framework, as an example. Similar results can be obtained on the other variants. From Figure~\ref{fig:abla_study}, two conclusions can be drawn. First, removing either $\mathcal{L}_\text{Orth}$ or LGWD from our framework causes an evident performance downgrade, validating the usefulness of these two developed techniques. Second, the ablated variants still outperform their underlying models. This further underpins the advantages offered by learning diverse filters with awareness of positional information.

\subsection{RQ3. Analysis on Diverse Filters}\label{sec:intep_filter_weight}
\begin{figure}[t]
     \centering
     \begin{subfigure}[b]{0.21\textwidth}
         \centering
         \includegraphics[width=0.9\columnwidth]{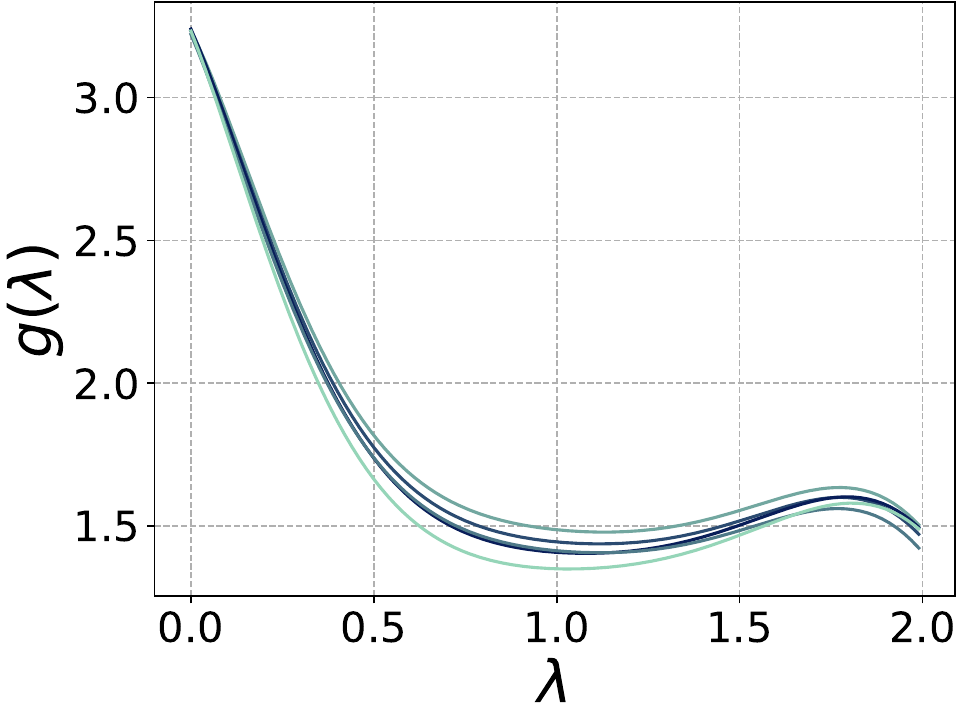}
         \caption{Citeseer}
         \Description{Results on Citeseer dataset.}
         \label{fig:LocFFuns_cite}
     \end{subfigure}
     \begin{subfigure}[b]{0.21\textwidth}
         \centering
         \includegraphics[width=0.9\columnwidth]{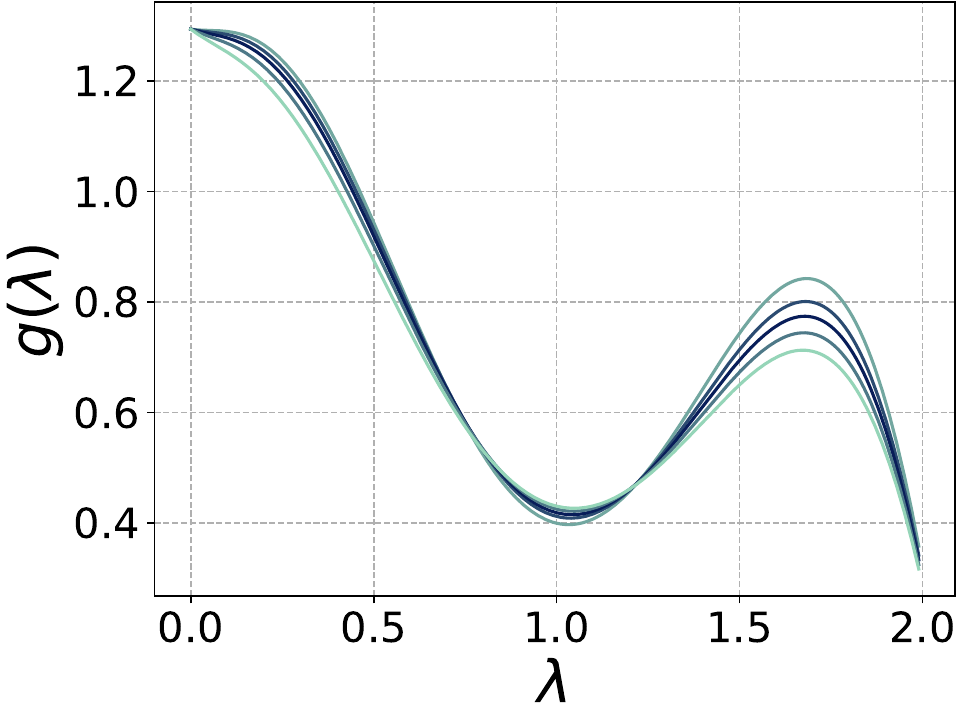}
         \caption{Photo}
         \Description{Results on Photo dataset.}
         \label{fig:LocFFuns_photo}
     \end{subfigure}
\caption{\small{Diverse filters on homophilic graphs, which are learned to be similar due to the intrinsic assortative linking patterns distributed uniformly on these networks. Our DSF presents one general framework which can be adaptive to different types of networks.}}
\Description{Figure 4. Fully described in the text.}
\label{fig:LocFFuns_hom}
\end{figure}

We now answer RQ3 by first plotting the diverse filter functions learned by our DSF with BernNet as the illustrative base model. Without loss of generality, we cluster the node-specific filter weights, i.e., $\{[\beta_{0,i},\beta_{1,i},...,\beta_{K,i}]^T | \forall v_i \in \mathcal{V}\}$, into five groups with k-means algorithm~\cite{jain1988algorithms}, and only plot the filters w.r.t. the representative centroids for better visualization. From Figure~\ref{fig:intep_visual} on heterophilc graphs, we observe a group of function curves showing similar overall shapes but different local aspects. This implies that the proposed DSF framework is able to grasp both conformal and disparate regional information on the graph. In addition, we also draw the diverse filters learned from homophilic graphs including Citeseer and Photo. These graph networks have assortative mixing patterns with homogeneous local structures. The learned filter functions produce almost identical curves fluctuating within a reasonable interval in Figure~\ref{fig:LocFFuns_hom}. It further shows  our DSF framework could work on different types of graphs. On the other hand, we present t-SNE~\cite{van2008visualizing} visualization of the node-specific filter weights. The color likeness reflects the corresponding similarity. From Figure~\ref{fig:nxFWs_cornell}, disparate regional patterns can be distinguished, and far-reaching nodes with conformal local subgraphs still learn similar filter weights. Besides, we notice the node in graph center displays a salient white color, obviously divergent from its neighborhood. This is because such vertex possesses the unique local context characterized by the densest graph neighborhood, and thereby deserves a special treatment. This phenomenon shows the flexibility of our DSF framework in dealing with complex or even unusual cases, instead of learning some strict relationships, e.g., nearby nodes ought to possess similar local structures (similar filter weights) and otherwise. We also present the visualization on graphs in relative larger scale, such as Chameleon network in Figure~\ref{fig:nxGraph_cham}. More results can be found in Appendix~\ref{apdix:add_expresults}. These analytical results demonstrate the strong interpretability of our DSF framework.

\begin{figure}[t]
\centering
\includegraphics[width=0.38\textwidth]{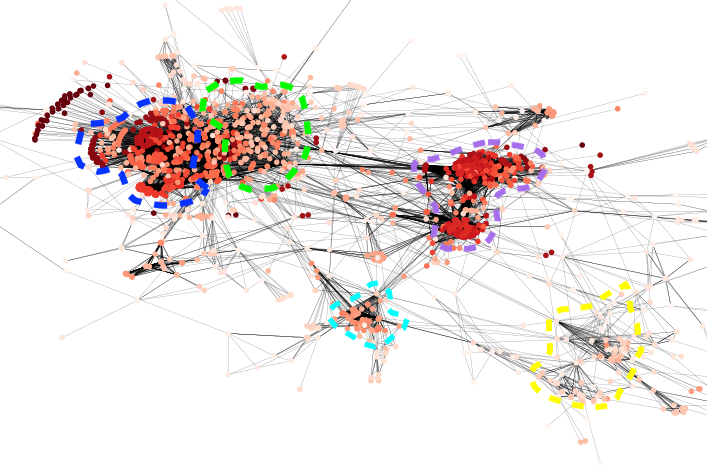}
\caption{\small{Visualization of node-specific filter weights on Chameleon dataset, where a few border nodes are cropped away for better picturing. We annotate the captured regional distinctions using irregular circles with different colors.}}
\Description{Figure 5. Fully described in the text.}
\label{fig:nxGraph_cham}
\end{figure}

\section{Conclusion}
This paper focuses on learning GNNs on complex graphs with regional \textit{heterogeneity} from the spectral perspective. 
We show that most existing spectral GNNs implicitly assume invariance between local networking patterns, and are restricted in the \textit{homogeneous} spectral filtering, thus limiting their performance. To this end, we propose a novel \textit{diverse} spectral filtering (DSF) framework generalizing  spectral GNNs to better exploit  rich and diverse local graph information.
Both theoretical and empirical investigations validate the effectiveness of our DSF framework and its enhanced interpretability by learning diverse filters.

\begin{acks}
The work was partially supported by the following: Jiangsu Science and Technology Programme (Natural Science Foundation of Jiangsu Province) under No. BE2020006-4; Key Program Special Fund in XJTLU under No. KSF-T-06.
\end{acks}

\bibliographystyle{ACM-Reference-Format}
\bibliography{ref}

\appendix

\section{Further Remarks about Related Work}\label{apdix:related_work_details}
Existing GNNs are often divided into spatial-based and spectral-based methods. The former is mainly built upon a message-passing framework~\cite{gilmer2017neural,balcilar2021breaking} where nodes exchange information  with their spatial neighbors. The latter is rooted in graph signal processing~\cite{shuman2013emerging} and spectral graph theory~\cite{chung1997spectral}, and are mostly divided into two categories. One class is spectral GNNs with fixed filters: GCN~\cite{kipf2017semi} truncates Chebyshev polynomials to a simple first-order, and works as a low-pass filter~\cite{wu2019simplifying}. APPNP~\cite{Klicpera2019PredictTP} constructs filter functions with personalized PageRank~\cite{page1999pagerank} to overcome the over-smoothing problem~\cite{li2018deeper}. GNN-LF/HF~\cite{zhu2021interpreting} is derived from the perspective of graph optimization to simulate low/high-pass filters. In the second class, spectral GNNs are mostly designed with trainable filters: ChebNet~\cite{defferrard2016convolutional} approximates the filtering operation with Chebyshev polynomials whose coefficients are learnable. AdaGNN~\cite{dong2021adagnn} learns adaptive filters to model each feature channel independently. GPR-GNN~\cite{chien2021adaptive} extends APPNP~\cite{Klicpera2019PredictTP} by directly parameterizing its filter weights and training them with gradient descent. ARMA~\cite{bianchi2021graph} takes rational filter functions while approximating them still with polynomials. BernNet~\cite{he2021bernnet} employs positive weight constraints in learning spectral filters with the Bernstein polynomial approximation. Wang and Zhang~\cite{Wang2022HowPA} further analyze the expressive power of spectral GNNs in a general form, and propose JacobiConv with an orthogonal polynomial basis and feature-wise filter learning.

\section{Proof of Proposition~\ref{prop:poly_subst}}
\begin{proof}
We denote  $f(x) = \sum_{k=0}^{K} \alpha_k P_k(x)$, and substituting $x = \xi \hat{x}$ gives us $f(\xi \hat{x}) = \sum_{k=0}^{K} \alpha_k P_k(\xi \hat{x})$. As the maximum order on variable $x$ is $K$, we can always express $f(\xi \hat{x})$ in a power series with new coefficient set $\{\omega_k\}_{k=0}^{K}$ where $\omega_k \in \mathbb{R}$, i.e., $f(\xi \hat{x}) = \sum_{k=0}^{K} \omega_k (\xi \hat{x})^k$. Moreover, since $\xi$ is a constant, we can view $f(\xi \hat{x})$ as a function of variable $\hat{x}$, i.e., $g(\hat{x}) = \sum_{k=0}^{K} (\omega_k \xi^k) \hat{x}^k$, Similarly, with the expressive power of polynomial basis $P_k(\cdot)$, we can always find a new coefficient set $\{\beta_k\}_{k=0}^{K}$ where $\beta_k \in \mathbb{R}$ making $g(\hat{x}) = \sum_{k=0}^{K} \beta_k P_k(\hat{x})$. Therefore, we have $f(x)=g(\hat{x})$ where these two functions are made up of the same polynomial basis $P_k; [0, 2] \to \mathbb{R}$ and two different coefficient sets $\{\alpha_k\}_{k=0}^{K}$ and $\{\beta_k\}_{k=0}^{K}$.
\end{proof}

\begin{table}[t]
\centering
\caption{\small{Statistics of real-world datasets, where $\star$ denotes large-scale graphs. Both $\mathcal{H}$~\cite{zhu2020beyond} and $\mathcal{H}_{\text{class}}$~\cite{lim2021new} (considering class-imbalance problem) measure graph homophily ratio from 0 to 1. Albeit the relative high value given by $\mathcal{H}=0.63$, Twitch-DE is essentially a heterophlic graph with class-imbalanced issue, as suggested by $\mathcal{H}_{\text{class}}=0.14$.}}
\label{tab:data_sta}
\setlength\tabcolsep{6pt}
\resizebox{00.47\textwidth}{!}{
\begin{tabular}{lcccccc}
\toprule[0.3pt]
Dataset   &\# Nodes &\# Edges &\# Features &\# Classes & $\mathcal{H}$ & $\mathcal{H}_{\text{class}}$ \\
\midrule[0.3pt]
% Chameleon &2,227       &36,101       &2,325          &5         &0.230   &0.062   \\
Chameleon &2,227       &36,101       &2,325          &5         &0.23   &0.06   \\
% Squirrel  &5,201       &217,073       &2,089          &5         &0.222   &0.025   \\
Squirrel  &5,201       &217,073       &2,089          &5         &0.22   &0.03   \\
% Wisconsin &251       &499       &1,703          &5         &0.178   &0.094   \\
Wisconsin &251       &499       &1,703          &5         &0.21   &0.09   \\
% Cornell   &183       &295       &1,703          &5         &0.296   &0.047   \\
Cornell   &183       &295       &1,703          &5         &0.30   &0.05   \\
% Texas     &183       &309       &1,703          &5         &0.061   &0.001   \\
Texas     &183       &309       &1,703          &5         &0.11   &0.00   \\
% Twtich-DE &9,498       &153,138       &2,514          &2         &0.632   &0.139   \\
Twitch-DE &9,498       &153,138   &2,545  &2         &0.63   &0.14   \\
\midrule[0.3pt]
% Cora      &2,708       &5,429       &1,433          &7         &0.810   &0.766   \\
Cora      &2,708       &5,429       &1,433          &7         &0.81   &0.77   \\
% Citeseer  &3,327       &4,732       &3,703          &6         &0.735   &0.627   \\
Citeseer  &3,327       &4,732       &3,703          &6         &0.74   &0.63   \\
% Pubmed$^\star$    &19,717       &44,324       &500       &5       &0.802       &0.664       \\
Pubmed$^\star$    &19,717       &44,338       &500       &3       &0.80       &0.66       \\
Computers$^\star$ &13,752       &245,861       &767          &10         &0.78   &0.70   \\
Photo     &7,650       &119,081       &745          &8         &0.83   &0.77   \\
\bottomrule[0.3pt]
\end{tabular}
}
\end{table}

\begin{algorithm}[t]
\caption{Framework of \textit{diverse} spectral filtering}
\label{alg:dsf_frame}
\begin{algorithmic}[1]
\REQUIRE
Node set: $\mathcal{V}$, Laplacian matrix: $\hat{\mathbf{L}}$, raw node content and positional features: $\mathbf{X} \in \mathbb{R}^{N \times f}, \mathbf{X}_p \in \mathbb{R}^{N \times f_p}$, polynomial basis: $P_k(\cdot)$, hyper-parameters: $K, \eta_1, \eta_2, \lambda_\text{Orth}$, ground truth labels for training: $\{\mathbf{y}_i \in \mathbb{R}^C| \forall v_i \in \mathcal{V}_\text{trn}\}$, activation function in Eq.~\ref{eq:pe_to_beta}: $\sigma_p(\cdot)$, and DSF-mode: $\phi \in \{\text{I}, \text{R}\}$.
\ENSURE 
$\mathbf{W}_x \in \mathbb{R}^{f \times d}$, $\mathbf{b}_x \in \mathbb{R}^d$, 
$\mathbf{W}_p \in \mathbb{R}^{f_p \times d}$, $\mathbf{b}_p \in \mathbb{R}^d$,
$\mathbf{W} \in \mathbb{R}^{d \times d}$, 
$\mathbf{W}_{F} \in \mathbb{R}^{d \times C}, \mathbf{b}_{F} \in \mathbb{R}^C$, 
$\{\mathbf{W}^{(k)} \in \mathbb{R}^{d}, \mathbf{b}^{(k)} \in \mathbb{R}|k=0,1,...,K\}$, and $\{\gamma_k \in \mathbb{R} | \forall k = 0,1...,K\}$.
% \STATE // \textit{Make an adjustment corresponding to different DSF-modes}.
\STATE Set $\eta_2 = 0$ if $\phi$ is R
\STATE // \textit{Projection into latent space}.
% \STATE $\mathbf{X}_i \gets \text{ReLU}(\mathbf{W}_x^T\:\mathbf{X}_i + \mathbf{b}_x)$, $\mathbf{P}^{(0)}_i \gets \text{Tanh}(\mathbf{W}_p^T\:{\mathbf{X}_p}_i + \mathbf{b}_b)$ for all $v_i \in \mathcal{V}$.
\STATE $\mathbf{X}_i \gets \text{ReLU}(\mathbf{W}_x^T\:\mathbf{X}_i + \mathbf{b}_x)$ for all $v_i \in \mathcal{V}$.
\STATE $\mathbf{P}^{(0)}_i \gets \text{Tanh}(\mathbf{W}_p^T\:{\mathbf{X}_p}_i + \mathbf{b}_b)$ for all $v_i \in \mathcal{V}$.
\STATE // \textit{enabled only for training}.
\STATE $\mathbf{X} \gets \text{Dropout}(\mathbf{X})$, $\mathbf{P}^{(0)} \gets \text{Dropout}(\mathbf{P}^{(0)})$. 

\STATE // \textit{Initialization}.
% \FOR{$v_i \in \mathcal{V}$}
%         \STATE $\theta_{0, i} \gets \sigma_p(\mathbf{W}^{(0)}^T \mathbf{P}_i^{(0)} + \mathbf{b}^{(0)})$
%         \STATE $\beta_{0, i} \gets \gamma_{k} \theta_{0, i}$.
% \ENDFOR
\STATE $\beta_{0, i} \gets \gamma_{0} \sigma_p({\mathbf{W}^{(0)}}^T \mathbf{P}_i^{(0)} + \mathbf{b}^{(0)})$ for all $v_i \in \mathcal{V}$.

\STATE $\mathbf{Z}^{(0)} \gets \text{diag}(\beta_{0, 1}, \beta_{0, 2}, ..., \beta_{0, N}) P_0(\hat{\mathbf{L}}) \mathbf{X}$.
\STATE // \textit{Iterate polynomial orders while updating node positions}.
\FOR{$k=1,2,...,K$}
    \STATE // \textit{Update node positional features}.
    \STATE $\mathbf{P}^{(k)} \gets \mathbf{P}^{(k-1)}$ using Eq.~\ref{eq:pe_iter_orth} with $\eta_1,\eta_2,\mathbf{W}$.
    
    \STATE // \textit{Update node hidden states}.
    % \FOR{$v_i \in \mathcal{V}$}
    %     \STATE $\theta_{k, i} \gets \sigma_p({\mathbf{W}^{(k)}}^T \mathbf{P}_i^{(k)} + \mathbf{b}^{(k)})$
    %     \STATE $\beta_{k, i} \gets \gamma_{k} \theta_{k, i}$.
    % \ENDFOR
    \STATE $\beta_{k, i} \gets \gamma_{k} \sigma_p({\mathbf{W}^{(k)}}^T \mathbf{P}_i^{(k)} + \mathbf{b}^{(k)})$ for all $v_i \in \mathcal{V}$.
    
    \STATE $\mathbf{Z}^{(k)} \gets \mathbf{Z}^{(k-1)} + \text{diag}(\beta_{k, 1}, \beta_{k, 2}, ..., \beta_{k, N}) P_k(\hat{\mathbf{L}}) \mathbf{X}$.
\ENDFOR
\STATE $\hat{\mathbf{y}}_i = \text{softmax}(\mathbf{W}_F^T\mathbf{Z}_{[i,:]}^{(K)} + \mathbf{b}_F), \forall v_i \in \mathcal{V}$. // \textit{Prediction}.
\STATE $\mathcal{L}_{\text{task}} = -\frac{1}{|\mathcal{V}_{\text{trn}}|}\sum_{v_i \in \mathcal{V}_{\text{trn}}}\mathbf{y}_i^T\log(\hat{\mathbf{y}}_i)$. // \textit{Training}.
\IF{$\phi$ is R}
    \STATE Minimize $\mathcal{L}_{\text{task}} + \lambda_\text{Orth}\mathcal{L}_{\text{Orth}}$ with $\mathcal{L}_{\text{Orth}}$ computed in Eq.~\ref{eq:orth_reg}.
\ELSE
    \STATE Minimize $\mathcal{L}_{\text{task}}$.
\ENDIF
\end{algorithmic}
\end{algorithm}

\section{Implementation Details}\label{apdix:model_details}
The overall pipeline of the proposed DSF framework is detailed in Algorithm~\ref{alg:dsf_frame}. 
We update node positional features and hidden states in an iterative and parallel scheme (see the lines 11-17). Noting in~\cite{dwivedi2022graph}, similar approaches can be found. In our experiments, we showcase it over three SOTA baselines including GPR-GNN, BernNet, and JacobiConv. A comprehensive summary of their designed trainable spectral filters can be found in~\cite{Wang2022HowPA}. We also provide other important implementation details in the following.

\begin{figure*}[t]
     \centering
    \begin{subfigure}[b]{0.459\textwidth}
         \centering
         \includegraphics[width=1\columnwidth]{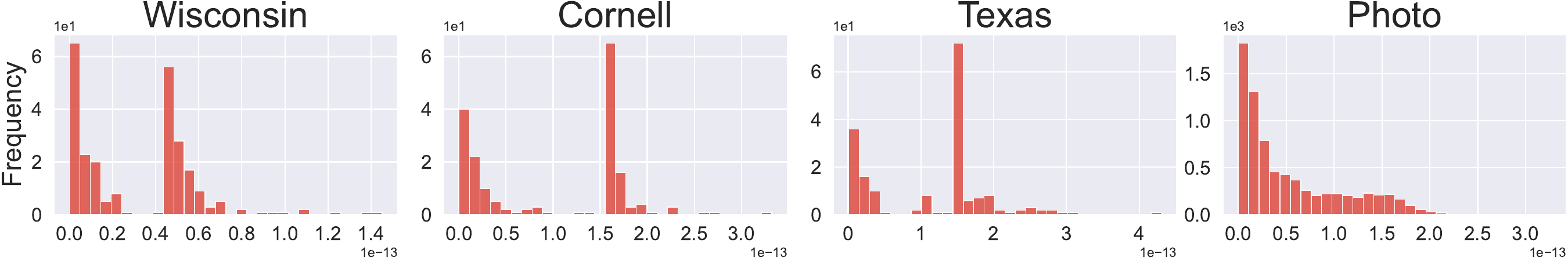}
         \caption{Local Graph Frequency (low-frequency)}
         \Description{Histograms of Local Graph Frequency on eigenvectors with low-frequency w.r.t. Wisconsin, Cornell, Texas, and Photo datasets.}
         \label{fig:lowFreqHist}
     \end{subfigure}
     \begin{subfigure}[b]{0.459\textwidth}
         \centering
         \includegraphics[width=1\columnwidth]{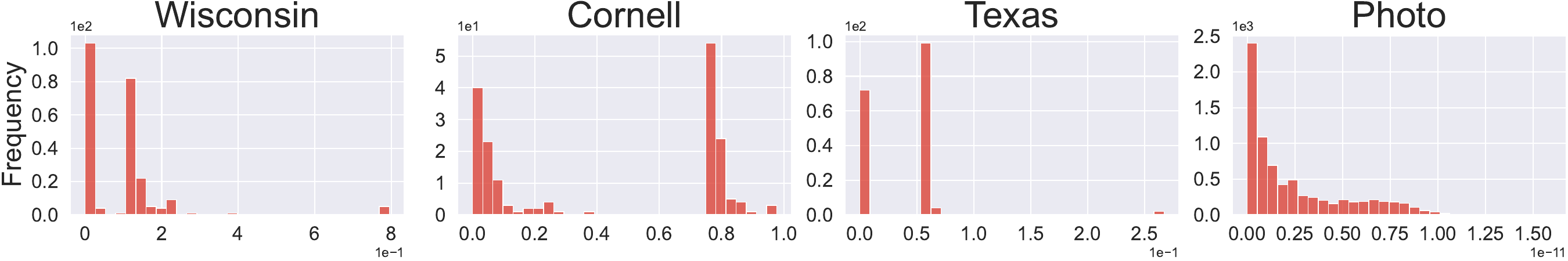}
         \caption{Local Graph Frequency (high-frequency)}
         \Description{Histograms of Local Graph Frequency on eigenvectors with high-frequency w.r.t. Wisconsin, Cornell, Texas, and Photo datasets.}
         \label{fig:highFreqHist}
     \end{subfigure}
     \begin{subfigure}[b]{0.459\textwidth}
         \centering
         \includegraphics[width=1\columnwidth]{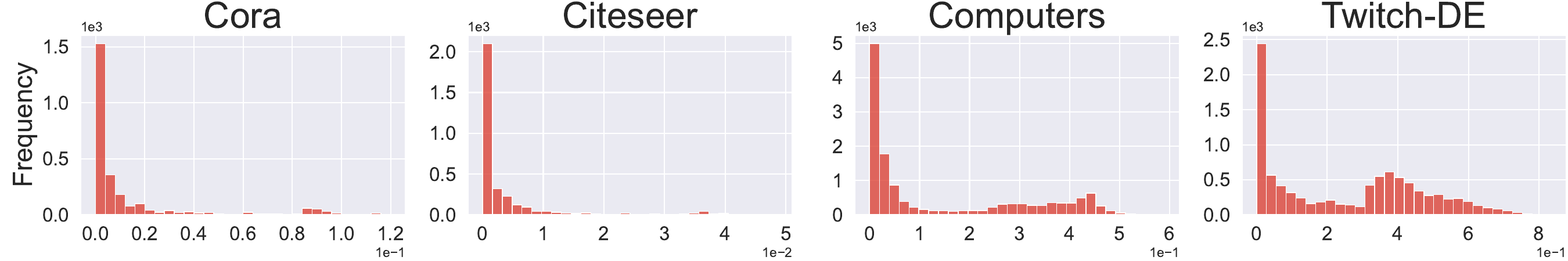}
         \caption{Local Graph Frequency (middle-frequency)}
         \Description{Histograms of Local Graph Frequency on eigenvectors with middle-frequency w.r.t. Cora, Citeseer, Computers, and Twitch-DE datasets.}
         \label{fig:supp__Ufreq_LocDiv}
     \end{subfigure}
     \begin{subfigure}[b]{0.459\textwidth}
         \centering
         \includegraphics[width=1\columnwidth]{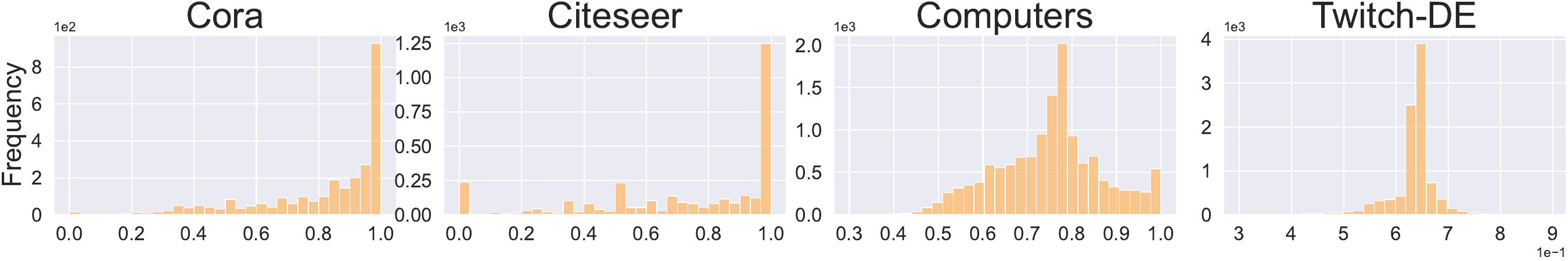}
         \caption{Local Label Homophily}
         \Description{Histograms of Local Label Homophily w.r.t. Cora, Citeseer, Computers, and Twitch-DE datasets.}
         \label{fig:supp__Hom_LocDiv}
     \end{subfigure}
\caption{Additional distributions of two essential graph properties (Better viewed in color). (a), (b), and (c) respectively show the statistics of local graph frequency using eigenvectors with low-, high-, and middle-frequency.}
\Description{Figure 6. Fully described in the text.}
\label{fig:supp__divGraphPattern}
\end{figure*}

\subsection{Base Model Information}
\textbf{GPR-GNN~\cite{chien2021adaptive} as backbone}: The authors experiment GPR-GNN with several initializing strategies on $\{\alpha_k\}^{K}_{k=0}$ and take the optimal one for the final evaluation. To take advantage of this, we adopt the same strategy to initialize our $\{\gamma_k\}^{K}_{k=0}$ before training. We call the resulted variants DSF-GPR-I and DSF-GPR-R.

\textbf{BernNet~\cite{he2021bernnet} as backbone}: As stated in the original paper, $\alpha_k$ is constrained to be non-negative. To follow up, we apply the same limit to our $\beta_{k,i} = \gamma_k \theta_{k,i}$ by making $\gamma_i \gets \text{ReLU}(\gamma_i)$ and taking $\sigma_p$ as a sigmoid function in Eq~\ref{eq:pe_to_beta} to restrict $\theta_{k,i}$ within $(0, 1)$. The resulted models are named as DSF-Bern-I and DSF-Bern-R.

\textbf{JacobiConv~\cite{Wang2022HowPA} as backbone}: The authors leverage a technique called PCD, which decomposes the filter weight into multiple coefficients, such as $\alpha_k = \pi_k \prod_{s=1}^k \rho_s$. To deploy our DSF framework over JacobiConv, we make $\pi_k=\gamma_k$, and transform $\rho_s$ into $\rho_{s,i}$ which is learned using Eq.~\ref{eq:pe_to_beta}. The produced variants are finally referred to as DSF-Jacobi-I and DSF-Jacobi-R.

\subsection{Initialization on IPE}\label{apdix:pe_init}
The choice of initializing node positional embeddings $\mathbf{X}_p$ is important, which usually requires to be permutation-invariant and distance-sensitive. In this work, we leverage two popular and efficient methods. The first one is widely used and called Laplacian Positional Encoding~\cite{belkin2003laplacian} (LapPE). It basically takes the decomposed eigenvectors $\{\mathbf{u}_1,\mathbf{u}_2,...,\mathbf{u}_N\}$, and each node $v_i$ is assigned with $\mathbf{P}^\text{Lap}_i = [\mathbf{u}_{1,i},\mathbf{u}_{2,i},...,\mathbf{u}_{f_p,i}]^T \in \mathbb{R}^{f_p}$, where $f_p \ll N$ is the predefined feature number. The other approach has been recently proposed based on the random walk diffusion process, named RWPE~\cite{dwivedi2022graph}. It aims to moderate the sign ambiguity issue in LapPE, and is formulated as $\mathbf{P}^\text{RW}_i = [\mathbf{RW}_{i,i}^1,\mathbf{RW}_{i,i}^2,...,\mathbf{RW}_{i,i}^{d_f}]^T \in \mathbb{R}^{f_p}$, where $\mathbf{RW} = \mathbf{A}\mathbf{D}^{-1}$. To increase model capacity and efficiency, before starting iterative update, we map $\mathbf{X}_p$ into a latent space with dimension $d$  (see line 3 in Algorithm~\ref{alg:dsf_frame}).

\section{Additional Experimental Results}\label{apdix:add_expresults}
\begin{figure}[ht]
\centering
\begin{subfigure}[b]{0.235\textwidth}
    \centering
    \includegraphics[width=1.0\columnwidth]{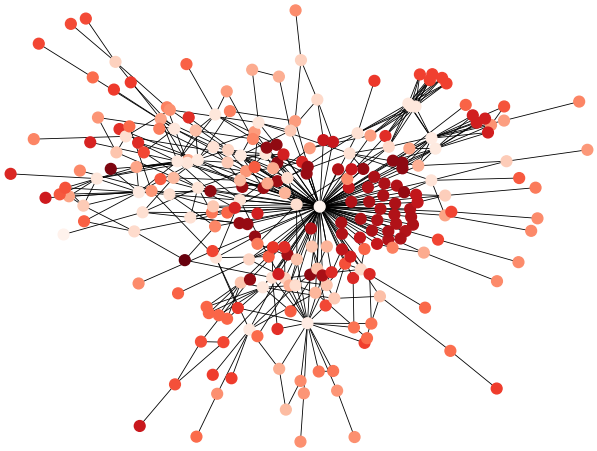}
    \caption{Wisconsin}
    \Description{Results on Wisconsin dataset.}
    \label{fig:nxGraph_wis}
\end{subfigure}
\hfill
\begin{subfigure}[b]{0.235\textwidth}
    \centering
    \includegraphics[width=1.0\columnwidth]{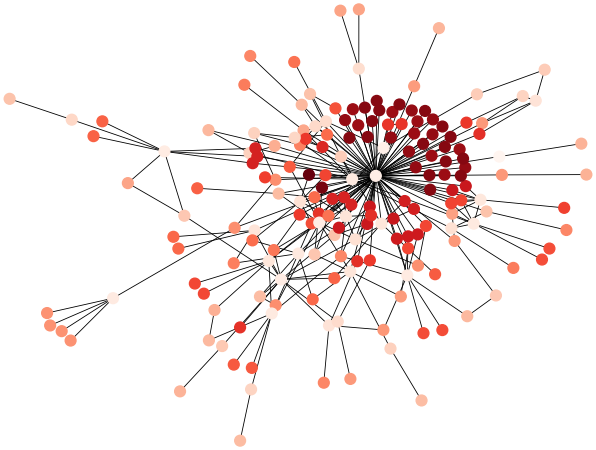}
    \caption{Texas}
    \Description{Results on Texas dataset.}
    \label{fig:nxGraph_texas}
\end{subfigure}
\vfill
\begin{subfigure}[b]{0.47\textwidth}
    \centering
    \includegraphics[width=1.0\columnwidth]{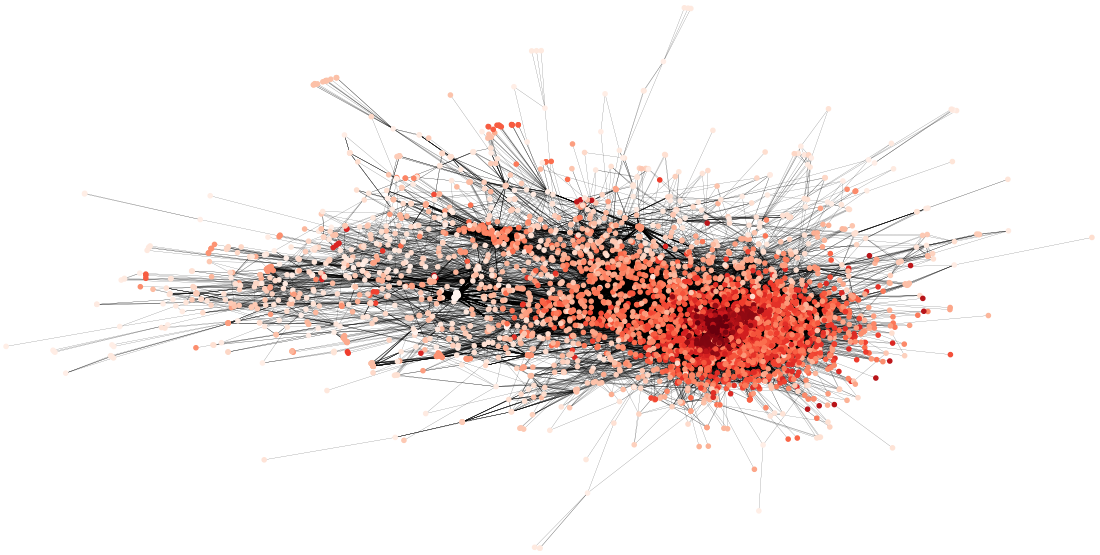}
    \caption{Squirrel}
    \Description{Results on Squirrel dataset.}
    \label{fig:nxGraph_sq}
\end{subfigure}
\caption{Additional visualization of node-specific filter weights.}
\Description{Figure 7. Fully described in the text.}
\label{fig:tsne_graph_fws}
\end{figure}

We also investigate the model performance in case of limited supervision. To do so, we follow the sparse splitting~\cite{chien2021adaptive} on homophilic graphs, i.e, 2.5\%/2.5\%/95\% for training/validation/testing. Table~\ref{tab:sparseNC} lists the classification accuracies (\%), where noticeable improvements are made by our DSF framework upon GPR-GNN albeit limited supervision. Figure~\ref{fig:tsne_graph_fws} provides more results on the visualization of node-specific filter weights. As the same type of networks, Wisconsin and Texas yield similar pictures to Cornell in the main text. For Squirrel dataset, we can see a gradual shift on the color depth of nodes from graph center to the border, which coincides with our conclusions in the main text about our DSF framework capturing regional heterogeneity.

\begin{table}[ht]
\caption{Classification accuracies (\%) on homophilic graphs with sparse splits.}
\Description{Our best variant makes noticeable improvements upon the base model by margin of 1.89, 1.38, 1.03, 1.97, and 0.64 percentage on Cora, Citeseer, Pubmed, Computers, and Photo datasets, respectively.}
\label{tab:sparseNC}
\setlength\tabcolsep{2pt}
\renewcommand{\arraystretch}{0.85}
\resizebox{0.48\textwidth}{!}{
\begin{tabular}{lcccccc}
\toprule[0.3pt]
\textbf{Datasets} & \textbf{Cora} & \textbf{Citeseer} & \textbf{Pubmed} & \textbf{Computers} & \textbf{Photo} \\
\midrule[0.3pt]
GPR-GNN
&76.05$\pm$0.48     &65.39$\pm$0.44   &83.30$\pm$0.24   &85.68$\pm$0.16    &91.88$\pm$0.18     \\
DSF-GPR-I
&77.75$\pm$0.42    &\textbf{66.83$\pm$0.37}   &83.74$\pm$0.15  &86.94$\pm$0.16    &92.39$\pm$0.14     \\
DSF-GPR-R
&\textbf{77.94$\pm$0.48}    &66.77$\pm$0.34    &\textbf{84.33$\pm$0.13}     &\textbf{87.65$\pm$0.15}    &\textbf{92.52$\pm$0.12}     \\
\midrule[0.3pt]
Our Improv.
&1.89\%     &1.38\%     &1.03\%        &1.97\%     &0.64\%      \\
\bottomrule[0.3pt]
\end{tabular}
}
\end{table}

\end{document}